%% file: paper.tex
\documentclass[]{fairmeta}

\title{Think Smarter not Harder: Adaptive Reasoning with Inference Aware Optimization}

\author[1,2,*]{Zishun Yu}
\author[1]{Tengyu Xu}
\author[1]{Di Jin}
\author[1]{Karthik Abinav Sankararaman}
\author[1]{Yun He}
\author[1]{Wenxuan Zhou}
\author[1]{Zhouhao Zeng}
\author[1]{Eryk Helenowski}
\author[1]{Chen Zhu}
\author[1]{Sinong Wang}
\author[1]{Hao Ma}
\author[1]{Han Fang}

\affiliation[1]{MetaAI}
\affiliation[2]{The University of Illinois Chicago}

\contribution[*]{Work done at Meta}

\abstract{

Solving mathematics problems has been an intriguing capability of large language models, and many efforts have been made to improve reasoning by extending reasoning length, such as through self-correction and extensive long chain-of-thoughts. While promising in problem-solving, advanced long reasoning chain models exhibit an undesired single-modal behavior, where trivial questions require unnecessarily tedious long chains of thought. In this work, we propose a way to allow models to be aware of inference budgets by formulating it as utility maximization with respect to an inference budget constraint, hence naming our algorithm Inference Budget-Constrained Policy Optimization (IBPO). In a nutshell, models fine-tuned through IBPO learn to ``understand'' the difficulty of queries and allocate inference budgets to harder ones. With different inference budgets, our best models are able to have a $4.14$\% and $5.74$\% absolute improvement ($8.08$\% and $11.2$\% relative improvement) on MATH500 using $2.16$x and $4.32$x inference budgets respectively, relative to LLaMA3.1 8B Instruct. These improvements are approximately $2$x those of self-consistency under the same budgets.
}

\date{\today}
\correspondence{Zishun Yu at \email{zyu32@uic.edu}}

\usepackage{tikz}
\usetikzlibrary{shapes.geometric, arrows.meta, positioning}
\usepackage{tcolorbox}

\usepackage{threeparttable}
\usepackage{graphicx}

\usepackage{multirow}

\usepackage{amsfonts}
\usepackage{amssymb, amsmath}
\usepackage{bbm}
\usepackage{mathtools}
\usepackage{wrapfig}
\usepackage{stmaryrd}
\usepackage{bm}

\usepackage{float}

\usepackage{algorithm}
\usepackage{algorithmic}
\usepackage{threeparttable}
\usepackage{listings}

\definecolor{codegreen}{rgb}{0,0.6,0}
\definecolor{codegray}{rgb}{0.5,0.5,0.5}
\definecolor{codepurple}{rgb}{0.58,0,0.82}
\definecolor{backcolour}{rgb}{0.95,0.95,0.92}
\definecolor{codedoc}{rgb}{0.13,0.55,0.13} 

\definecolor{RoyalBlue}{rgb}{0.2549, 0.4118, 0.95}
\definecolor{BrickRed}{rgb}{0.85, 0.3, 0.3294}

\lstdefinestyle{mystyle}{
    backgroundcolor=\color{backcolour},   
    commentstyle=\color{codegreen},
    keywordstyle=\color{magenta},
    numberstyle=\tiny\color{codegray},
    stringstyle=\color{codepurple},
    basicstyle=\ttfamily\footnotesize,
    breakatwhitespace=false,         
    breaklines=true,                 
    captionpos=b,                    
    keepspaces=true,                 
    numbers=left,                    
    numbersep=5pt,                  
    showspaces=false,                
    showstringspaces=false,
    showtabs=false,                  
    tabsize=2,
    morecomment=[s][\color{codedoc}]{"""}{"""}, 
}
\newcommand{\E}{\mathbb{E}}
\newcommand{\xmat}{\mathbf{X}}
\newcommand{\ymat}{\mathbf{Y}}
\newcommand{\rmat}{\mathbf{R}}

\newcommand{\xcal}{\mathcal{X}}
\newcommand{\ycal}{\mathcal{Y}}
\newcommand{\vcal}{\mathcal{V}}
\newcommand{\D}{\mathbb{D}}

\newcommand{\g}{\mathcal{G}}
\newcommand{\J}{\mathcal{J}}

\newcommand{\ind}{\mathbbm{1}}

\newcommand{\KL}{\mathbb{KL}}

\DeclareMathOperator*{\argmax}{arg\,max}
\DeclareMathOperator*{\argmin}{arg\,min}

\begin{document}

\maketitle

\input{1.intro}

\input{2.design}
\input{3.implementation}

\input{4.construction}

\input{5.performance}

\input{6.conclusion}

\bibliographystyle{assets/plainnat}
\bibliography{ref}

\newpage
\appendix
\beginappendix

\input{appendix/apx_responses}

\input{appendix/apx_accumulation}

\input{appendix/apx_solver}

\input{appendix/apx_data}

\input{appendix/hyperparams}

\input{appendix/apx_budget_following}

\end{document}

%% file: 1.intro.tex
\newtheorem{proposition}{Proposition}

\section{Introduction}
\label{section: intro}

Complex reasoning has been an intriguing ability of large language models (LLMs), with application in for example mathematical problem-solving~\citep{cobbe2021training, hendrycks2021measuring, lightman2023let} or coding~\citep{chen2021evaluating, austin2021program, hendrycks2021measuringcode}, which does not only require nature language comprehending but also logical and critical ``thinking''. An observation merged in the LLM reasoning literature is that longer reasoning traces often leads to improved reasoning soundness and correctness. The seminal work of chain-of-thought (CoT)~\citep{wei2022chain} is an excellent example of how enriching reasoning details, by decomposing reasoning traces into steps, improves its problem-solving capability. CoT has been considered a standard technique in reasoning, recent works extend CoT by allow LLMs to expand its reasoning steps, by for example CoT with more steps~\citep{jin2024impact} (as explicitly required by instruction), self-reflection/correction~\citep{madaan2024self, zelikman2022star, yan2024s, qu2024recursive}, multi-turn reasoning~\citep{kumar2024training} or multi-agent debate~\citep{liang2023encouraging, pham2023let} (as a heterogeneous case of multi-turn). It was conjectured that scaling the test-time compute or the reasoning length unleashes LLMs' potential for reasoning~\citep{snell2024scaling}, which has been empirically verified by recent hype of ultra-long reasoning models, such as OpenAI-o1~\citep{jaech2024openai} and DeepSeek-R1~\citep{deepseekai2025deepseekr1incentivizingreasoningcapability}. We'll later categorizes these type of responses as (standard) CoT responses, extended responses, and (ultra-)long responses, respectively, given the nature of their reasoning lengths.

\begin{wrapfigure}[10]{r}{0.45\textwidth}
    \vspace{-0em}
    \centering
    \includegraphics[width=\linewidth]{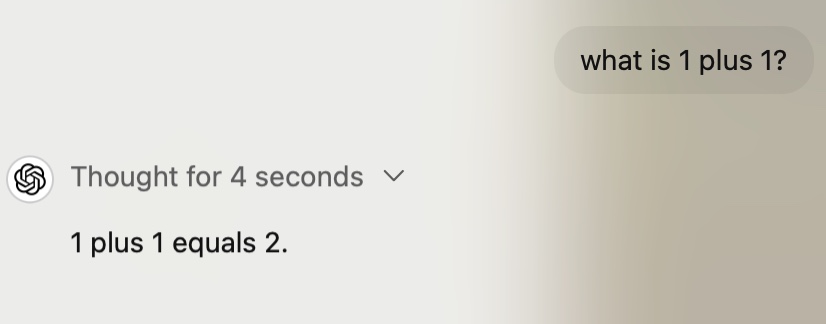}
    \caption{A long reasoning-chain model spent more than enough inference time on a trivial problem.}
    \label{fig:o1}
\end{wrapfigure}

While scaling reasoning length is promising in problem-solving, advanced long reasoning-chain models show an undesired uni-modal behavior that trivial questions may require unnecessarily tedious long reasoning trace, an example is shown in Figure~\ref{fig:o1}. 
This uni-modal behavior creates unnecessarily higher inference costs and increased carbon footprints~\citep{henderson2020towards, anthony2020carbontracker}. 
To partially address this, we study how to enable multi-modal behavior for reasoning models in a way the length of reasoning traces are automatically adjusted according to the hardness level of the queries. From the aspect of query-adaptive reasoning length, some heuristic methods~\citep[e.g.][]{aggarwal2023let, xu2024adaption, wang2024make} have 
been making effort towards better token efficiency, by which is meant better accuracy with (hopefully) less token overhead.
We take a reinforcement learning (RL) perspective, 
where the accuracy gain over the token overhead is nothing but a non-differentiable objective to be optimized. 
One could, for instance, take the negative response length or a metric of this sort as an intrinsic reward~\citep{chentanez2004intrinsically, pathak2017curiosity}. However, balancing the intrinsic and extrinsic (accuracy) rewards might also be challenging~\citep{liu2021decoupling}, and might be vulnerable to reward hacking~\citep{pan2022effects, skalse2022defining, karwowski2023goodhart}.

Instead explicit modeling the length of responses, we take a more abstract formulation, where we consider labeling each response $y$ with an unique group label $\g_i: i \in \llbracket G \rrbracket$ for total number of $G$ groups, so that the union of these disjoint groups exactly form the response space $\cup_i \g_i = \ycal$.
For example $\g$ could be the group of CoT responses (with standard length) or extended responses. We could then impose an density constraint on each or a set of groups, by caping $\E_{x\sim\mu}\E_{y\sim\pi(x)}[\ind_{\{y\in\g_i \}}] \leq q_i$ for some prompt distribution $\mu$ and some response distribution $\pi(x)$, induced from LLMs, conditioned on a prompt $x$.  
This naturally formulates a constrained RL~\citep{garcia2015comprehensive, altman2021constrained} problem. 
Also this group definition is motivated by the resource-allocation literature~\citep{chenery1956resource, ibaraki1988resource, karlin2003mathematical}, from the optimization and econometric communities, which have been later applied in many machine learning applications~\citep[e.g.][]{zemel2013learning, badanidiyuru2018bandits}. This generalization allows potential broader application of our algorithm as discussed in Section~\ref{sec:con}. We've now set our goal of this work:
\begin{tcolorbox} 
\centering
A constrained RL framework controlling how response groups $\{\g_i\}$ are distributed.
\end{tcolorbox}
Therefore, one could control how responses of different lengths (which are supposed to belong to different groups) are distributed.
Our rationale of  algorithm design is given in Section~\ref{sec:design}, derived from an optimization perspective but ended as a very simple generalization of iterative supervised fine-tuning (SFT) methods such as reward-ranking fine-tuning (RAFT)~\citep{dong2023raft} and rejection sampling fine-tuning (RFT)~\citep{ouyang2022training, touvron2023llama}, see details in Section~\ref{sec:algo}. Given the motivation of our algorithm, we call the resulted algorithm as Inference Budget-Constrained Policy Optimization (IBPO).

{\bf Paper structure.} In Section~\ref{sec:design}, we present the derivation of our algorithm from an optimization perspective, resulting in a simple weighted iterative SFT update. Section~\ref{sec:algo} provides further details on our practical implementation, including the choice of the base algorithm and the design of reward. Section~\ref{sec:sv} introduces the experimental settings used for empirical evaluation. The final empirical results of our IBPO are presented in Section~\ref{sec:exp}. 
And Section~\ref{sec:con} concludes our work with limitations, broader impact, and further discussions.

%% file: 2.design.tex
\section{Algorithm Design}\label{sec:design}

{\bf Problem setup.} To make the notation compact, we take the sequence-level notation (or the bandit notation), commonly used in LLMs~\citep{ziegler2019fine, rafailov2024direct}, especially in preference modeling, that suppresses the transition probabilities and intermediate rewards, and see a response as a whole. 
In particular, a policy $\pi : \xcal \to \Delta(\ycal)$ takes a prompt $x \in \xcal$ and draw a response $a_1 \circ a_2 \dots \circ a_T =: y \in \ycal$ from the produced probability simplex $\Delta(\ycal)$, where $\circ$ denotes concatenation, $a_i \in \vcal$ corresponds to the $i$-th token drawn from the vocabulary $\vcal$, and $T$ is the maximum length. 
A LLM is a parametric policy $\pi_\theta \in \Pi_\theta \subseteq \Pi$, where $\Pi_\theta$ and $\Pi$ are the parametric and non-parametric policy space respectively. 
Let $\J(\pi; \mu, r)$ or sometimes $\J(\pi)$ be a general objective function, defined by a prompt distribution $\mu \in \Delta(\xcal)$, and a bounded reward  function $r:\xcal \times \ycal \to [-R_\mathrm{max}, R_\mathrm{max}]$. Also, we define ${\mu}_{\Omega}$ as an empirical distribution induced from $\Omega$, a set of prompts.

As aforementioned, we define $G$ disjoint groups $\g_i$ such that $\cup_i \g_i = \ycal$ and $\g_i \cap \g_j = \emptyset$ for all $i \neq j$. Each response $y \in \ycal$ is attached to exactly one group, in the sense that $y \in \g_i$ for some $i$. 
In the context of LLM reasoning, without loss of generality, we consider two groups for brevity: $\g_\circ$ and $\g_+$, corresponding to regular-length CoT responses and extended responses (with low and high inference costs), respectively.
To conclude the formulation of constrained RL with resource allocation constraints, we could in general define the feasible set as a convex polytope, $\Phi_\g := \{\pi: \E_{x}\E_{y\sim\pi(x)}[\ind_{\{y\in \g_i \}}] \leq q_i \text{~for all $i$}\}$, that caps the total density mass of each group $\g_i$ by $q_i$. In our setting, we only need to cap the total mass of extended responses to optimize the inference efficiency, posing a half-space $\Phi_+ := \{\pi: \E_{x}\E_{y\sim\pi(x)} [\ind_{\{y\in \g_+ \}}] \leq q_+ \}$ for some $q_+ > 0$.

{\bf Background.} With the bandit setup, our setting draw a lot connection to the online learning literature, especially online/bandit convex optimization~\citep{hazan2016introduction, slivkins2019introduction}, although we optimize a fixed function the training data is however collected in an online fashion. In the sense of distributing resources across groups, it is connected to for example knapsack bandits~\citep{badanidiyuru2018bandits} and statistical parity~\citep{zemel2013learning}, aka group fairness. Although the definition of groups and optimization programs could be different. This allocation optimization point of view allows us to further extend our method to broader LLM applications. To the end of policy optimization with respect to constraints, common techniques include projection~\citep{zinkevich2003online, flaxman2004online, bubeck2015bandit, yang2020projection}, Lyapunov-based~\citep{chow2018lyapunov, cayci2022lyapunov}, or Lagrangian methods~\citep{ray2019benchmarking}.

{\bf Non-parametric space $\Pi$.} In our case, our goal is to solve:
\begin{align}
    \textstyle\max_{\pi_\theta \in \Pi_\theta} \J(\pi_\theta) \quad \text{s.t.}~\pi_\theta \in \Phi_+ \quad \text{where $\Pi_\theta$ is the parameterized policy space.}  \label{eq:original-opt}
\end{align}
Solving Eq.~\eqref{eq:original-opt} is however intractable due to the LLM parameterized policy space $\Pi_\theta$.
A common practice is alternating gradients between reward maximization and constraint satisfaction. For example in Lagrangian methods such as TRPO/PPO-Lagrangian~\citep{ray2019benchmarking}, one could do alternating update $\pi_\theta$ and the Lagrangian multiplier.
Another workaround is to first obtain a solution $\pi^\star$ in the non-parametric policy space $\Pi := \Delta(\xcal \times \ycal)$, aka tabular representations, and project $\pi^\star$ onto the parameteric one $\Pi_\theta$, as a technique used in many (constrained) RL works~\citep{peters2010relative, montgomery2016guided, zhang2020first}.

The advantage of working in the non-parametric space $\Pi$ is: solving $\max_{\pi\in\Pi} \J(\pi)$ s.t. $\pi\in\Phi_+$ is easy, on the conditions that (i) $\J(\pi)$ is concave in $\pi$ so that it is a convex program, and (ii) sampling and evaluating the reward function $r(\cdot, \cdot)$ and the cost indicator $\ind_{\{\cdot\}}$ are cheap. 
The condition (i) is sometimes true, for instances, bandit objective~\citep{slivkins2019introduction}; the LP formulation~\citep{manne1960linear, denardo1970linear, nachum2020reinforcement, nachum2019dualdice} of RL, and (relative) entropy regularized RL~\citep{ziebart2010modeling, haarnoja2017reinforcement, haarnoja2018soft} are often concave in occupancy measure $\rho$. 
However, in rare case condition (ii) holds, RL works~\citep{haarnoja2018soft, peng2019advantage, zhang2020first} often resort to value function approximation, making it is easy, for discrete action space, to evaluate $Q$-values (or alternatively advantages) for all actions for a specific state. 
It is then tractable to obtain closed-form solution (optimal w.r.t. the value/advantage approximations) in $\Pi$. Once an optimal policy $\pi^\star$ is found, one could then project it onto $\Pi_\theta$ through (reverse) information projection $\theta = \argmin_\theta \KL(\pi^\star \Vert \pi_\theta)$ (which is often done approximately by taking gradient steps).

{\bf Stochastic optimization.} With a LLM, it is obviously intractable to sample and evaluate the reward function $r(x, y)$ for all $(x, y) \in \xcal \times \ycal$, similarly for the cost indicator $\ind_{\{\cdot\}}$. 
To avoid the training of additional value models for LLMs~\citep{snell2022offline, yu2023beta}, which can create significant overhead in terms of memory usage, implementation complexity, and training stability, 
we consider a stochastic optimization. The stochastic counterpart as described in Eq.~\eqref{eq:empirical} solves an approximate $\hat{\pi}^\star$ using a manageable number of samples rather than directly solving for the global optimum $\pi^\star$, still, in the non-parametric space $\Pi$:
\begin{equation}
\begin{aligned}
    \hat{\pi}^\star(\xmat, \ymat) = &\textstyle\argmax_{\pi \in \Pi} \hat{\J}(\pi; \xmat, \ymat) := \frac{1}{nm}\sum_i^n\sum_j^m [\pi(y_{ij}|x_{i})r(x_i, y_{ij})] \\
    &~\text{s.t.}~\pi \in \hat{\Phi}_+(\xmat, \ymat) := \{\pi: \textstyle\sum_i\sum_j [\pi(y_{ij}|x_i) \left( \ind_{\{y_{ij} \in \g_+ \}} - q_+ \right)] \leq 0 \}
\end{aligned}\label{eq:empirical}
\end{equation}
where $\xmat \in \xcal^n$ is a vector of $n$ sampled prompts and $\ymat \in \ycal^{n \times m}$ is a matrix of responses, with $m$ responses for each of the $n$ prompts; we explicitly write the empirical objective $\hat{\J}$ with the conventional expected reward maximization for notational convenience, though alternative objectives are not restricted.

Since the empirical problem~\eqref{eq:empirical} is a convex program with relative small sample size, it is now manageable. 
Combing the aforementioned projection step, we could write the program as a bi-level optimization:
\begin{align}
    \textstyle\pi_\theta = \argmin_{\pi_\theta \in \Pi_\theta} \E_x \left[ \KL( \hat{\pi}^\star_{\xmat, \ymat_\theta} \Vert \pi_\theta)[x] \right] \quad \text{s.t.}~\hat{\pi}^\star_{\xmat, \ymat_\theta} \in \argmax_{\pi\in \Pi\cap \hat{\Phi}_+(\xmat, \ymat_\theta)} \hat{\J}(\pi; \xmat, \ymat_\theta)
\end{align}
where $\ymat_\theta \sim \pi_{\theta}(\xmat)$, and hence $\hat{\pi}^\star$ is indirectly a function of $\theta$.

{\bf Practical update.} 
For general bi-level optimization, iteratively solving the upper and lower-level problems by alternatively fixing one while optimizing the other could be expensive~\citep{zhang2024introduction}. 
We've already setup a manageable inner problem, making it easy to solve for example using convex solvers. One could therefore do iterative gradient updates on the upper level while directly solving the lower-level at each iteration:
\begin{align}
    \textstyle\theta' = \theta - \alpha \nabla_\theta {\E}_{x\sim\mu_\xmat} \left[ \KL(  \hat{\pi}^\star_{\xmat, \ymat_\theta} \Vert \pi_\theta)[x] \right] \quad \text{s.t.}~\hat{\pi}^\star_{\xmat, \ymat_\theta} \in \argmax_{\pi\in \Pi\cap \hat{\Phi}_+(\xmat, \ymat_\theta)} \hat{\J}(\pi; \xmat, \ymat_\theta) 
\end{align}
where $\theta$ and $\theta'$ are the parameters of current and next iteration, respectively; and the projection step is also done as stochastic optimization with samples $(\xmat, \ymat_\theta)$ of the current iteration.

Note that $\hat{\pi}^\star$ is indirectly a function of $\theta$ through the samples $(\xmat, \ymat_\theta)$. The gradient $\nabla_\theta \KL(  \hat{\pi}^\star_{\xmat, \ymat_\theta} \Vert \pi_\theta)$ hence requires differentiation through $\hat{\pi}^\star_{\xmat, \ymat_\theta}$, meaning differentiate through an $\argmax$ operator, which can in principle be achieved through implicit differentiation~\citep{amos2017optnet, lorraine2020optimizing}.
However, to avoid additional implementation and computation overhead, 
instead we use the semi-gradient ${\nabla}_\theta \KL( \textsc{sg}\{ \hat{\pi}^\star_{\xmat, \ymat_\theta} \} \Vert \pi_\theta)$, where $\textsc{sg}\{\cdot\}$ is a stop gradient operator.
This stop-gradient trick is quite common in many ML applications~\citep{sutton2018reinforcement, foerster2018dice, chen2021exploring}, leading to the update:
\begin{align}
    \underbrace{\textstyle\theta' = \theta - \alpha \nabla_\theta {\E}_{x\sim\mu_\xmat} \left[ \KL(  \textsc{sg}\{\hat{\pi}^\star_{\xmat, \ymat_\theta}\} \Vert \pi_\theta)[x]\right] }_\text{approximate projection / weighted SFT}
    \quad \text{s.t.}~\underbrace{\hat{\pi}^\star_{\xmat, \ymat_\theta} \in \textstyle\argmax_{\pi\in \Pi\cap \hat{\Phi}_+(\xmat, \ymat_\theta)} \hat{\J}(\pi; \xmat, \ymat_\theta) }_\text{optimization for weight}\label{eq:semi}
\end{align}

The semi-gradient ${\nabla}_\theta \KL( \textsc{sg} \{\hat{\pi}^\star_{\xmat, \ymat_\theta} \} \Vert \pi_\theta)[x_i] = 
-\sum_{j}
\hat{\pi}^\star_{\xmat, \ymat_\theta}(y_{ij} | x_i)
\frac{\partial}{\partial \theta}
\log \pi_\theta(y_{ij} | x_i)$ is {\color{BrickRed} a weighted SFT update (via $\hat{\pi}^\star$)}. This observation creates an extremely simple update with negligible implementation overhead.

{\bf Discussion.} The update rule ended up aligning many iterative weighted SFT algorithms, such as RAFT~\citep{dong2023raft} and RFT~\citep{ouyang2022training, touvron2023llama}.
In hindsight, our algorithm is motivated by the observation that these extremely successful algorithms can be interpreted as projecting empirical solutions onto a parametric space. 
Consequently, it is reasonable to use the empirical estimate in Eq.~\eqref{eq:empirical}, as RAFT and RFT have demonstrated strong practical performance despite the inherent bias introduced by the non-linearity of these estimations.
Since it is essentially generalizes SFT by re-weighting a sample pair $(x_i, y_{ij})$ by $\hat{\pi}^\star(y_{ij}|x_i)$, at each iteration based on the solution of an optimization problem $\hat{\pi}^\star$.
We create a Table~\ref{tab:sft-algos} outlines the corresponding components for some iterative SFT methods from this optimization point-of-view.

\begin{table}[!tbp]
    \centering
    \caption{SFT methods from our optimization point-of-view. $\mathbb{XE}$ denotes cross-entropy loss. RM stands for reward model, binary reward indicates being correct or incorrect. $\hat{\pi}^\star$ are (unnormalized) weights for the subsequent $\mathbb{XE}$ update.}
    \label{tab:sft-algos}
    \begin{tabular}{l | cccc}
        \toprule
        \textbf{ALGO} & loss $\mathcal{L}$ & reward $r$ & feasible set $\Phi$ &  weight/acceptance $\hat{\pi}^\star$ \\
        \midrule
        {(Iterative) SFT} & $\mathbb{XE}$ & constant & $\Pi$ & constant \\
        
        {RFT} & weighted $\mathbb{XE}$ & binary & $\Pi$ & $\ind_{\{r(x, y) = 1\}}$ \\
        
        RAFT (Best-of-$N$) &  weighted $\mathbb{XE}$ & RM & $\Pi$ &  $\ind_{\{r(x_i, y) = \max_{j}r(x_i, y_{ij}) \}}$ \\
        \midrule
        Ours &  weighted $\mathbb{XE}$ & Section~\ref{sec:algo} &  Eq.~\eqref{eq:empirical} & Eq.~\eqref{eq:empirical} \\
        \bottomrule
    \end{tabular}
\end{table}
The optimization problems for RFT and RAFT are trivial, as they assign $\hat{\pi}^\star(y|x) = 1$ to accepted responses and to the response with the highest reward model score, respectively. 
Formulating these methods as optimization does not offer much advantages. However, this perspective provides flexibility for future work to extend our framework, allowing for different feasible sets and weighting schemes tailored to specific application needs.

%% file: 3.implementation.tex
\section{Practical Implementation}\label{sec:algo}

Yet, as we are working in an algorithm-agnostic fashion, we are now ready to select a specific RL algorithm, define its corresponding objective $\J$, and specify an appropriate reward function $r$.

{\bf Reward function.} Since we are working on mathematical problem-solving, a ground-truth reward could be obtained through string matching~\citep{cobbe2021training, hendrycks2021measuring} of the model’s solution against the ground truth solution, yielding a binary reward function $r_{\mathrm{match}}: \xcal\times\ycal \to \{0, 1\}$ that indicates correctness. On top of the binary reward, we define our reward $r_\mathrm{\Delta}$ as the reward margin.
To formally construct the margin, we first define the expected reward of a set $\g$ such that $\bar{r}_{\pi}(x, \g) := \E_{y\sim\pi}[r_\mathrm{match}(x, y) \mid y\in\g]$. We then define the reward margin $r_\mathrm{\Delta}$ as the reward advantage of a group $\g$ against all other groups $\ycal \setminus \g$:
\begin{align}
    r_\Delta(x, y \in \g) := \bar{r}_\pi(x, \g) - \bar{r}_\pi(x, \ycal\setminus \g)
\end{align}
In our case of two groups, we have $r_\Delta(x, y \in \g_+) = \bar{r}_\pi(x, \g_+) - \bar{r}_\pi(x, \g_\circ)$ and similarly for $\g_\circ$. 
While $r_\Delta$ might appear odd for not addressing the correctness of individual  $y$, this is handled subsequently.

{\bf RL objective.} For the learning algorithm, our choice is Constraint Generative Policy Optimization~(CGPO)~\citep{xu2024perfect}, which was originally designed for multi-objective constrained optimization of LLMs. 
The choice is driven by implementation considerations: CGPO’s modular constraint-handling design makes it straightforward to incorporate additional constraints, such as the group density constraint in our case.
CGPO in a nutshell can be viewed as a generalized Best-of-$N$ (BoN), though depending on the specific CGPO settings.
It operates by defining a feasible set $\Xi$ over the sample space $\xcal\times\ycal$, which are designed to capture constraint adherence for, e.g. correctness, factuality, and safety~\citep{xu2024perfect}. In short, $(x, y)\notin\Xi$ will be rejected.
In the context of math reasoning, the objective of CGPO can be summarized as:
\begin{equation}
\begin{aligned}
    \max_\pi \E_x \E_{y\sim\pi} [r(x, y)] \quad &\text{s.t.}~{\textstyle\sum_y[\pi(y|x)\ind_{\{y\in\Xi_x\}}] \geq 1}~\text{for all $x$} \\ 
    &\text{where}~\Xi_x := \{\underbrace{y: r_\mathrm{match}(x, y) = 1}_\text{correctness} \} \cap \{\underbrace{y: \hat{\KL}(y; x, \pi_\mathrm{ref}) \leq \KL_\mathrm{max}}_\text{empirical KL} \}\label{eq:cgpo}
\end{aligned}
\end{equation}
This objective essentially optimizes reward over the feasible sets $\Xi_x$ such that feasible  response $y$ is correct and within an KL range of $\KL_\mathrm{max}$, where the KL constraint is measured using point estimate of the (forward) KL defined as $\hat{\KL}(y;x,\pi_\mathrm{ref}):= \log\pi(y|x) - \log\pi_\mathrm{ref}(y|x)$.

\begin{table}[!tbp]
    \centering
    \begin{threeparttable}
    \caption{A concrete example of $\textsc{opt}_\mathrm{IuB}$ compared to $\textsc{opt}_\mathrm{cgpo}$-Eq.~\eqref{eq:cgpo}- with \underline{tie broken randomly}, resulting in potentially non-unique $\hat{\pi}^\star$. This table shows several such $\hat{\pi}^\star$ solutions but not all. KL constraint is omitted for brevity. Given the rewards defined below, we have $\bar{r}(x_1, \g_\circ) = \bar{r}(x_1, \g_+) = \bar{r}(x_2, \g_+) = 1$ and $\bar{r}(x_2, \g_\circ) = 0.5 $. Suppose the density constraint $q_+ = 0.5$, allowing at most $50$\% of accepted responses are extended $y\in\g_+$. For a solution matrix $\hat{\pi}^\star$, $1$ and $0$ represent accepted and rejected response, respectively.}
    \label{tab:iub-example}
    \begin{tabular}{l l | cc | c | c | c }
        \toprule
        \textbf{prompt} & \textbf{responses} & $r(x, y)$ & $r_\Delta(x, y)$ & 
        $\hat{\pi}^\star_1 = \textsc{opt}_\mathrm{cgpo}$\tnote{$\dagger$} & 
        $\hat{\pi}^\star_2 = \textsc{opt}_\mathrm{cgpo}$\tnote{$\ddagger$} & $\hat{\pi}^\star = \textsc{opt}_\mathrm{IuB}$\tnote{$\sharp$} \\ 
        \midrule
        \multirow{2}{*}{$x_1$ (easy)} & $y_{11}, y_{12} \in \g_\circ$ & \multirow{2}{*}{
            $\begin{pmatrix}
                1 & 1 \\
                1 & 1 \\
            \end{pmatrix}$
        } & 0 & \multirow{2}{*}{
            $\begin{pmatrix}
                0 & 0 \\
                1 & 0 \\
            \end{pmatrix}$
        } & \multirow{2}{*}{
            $\begin{pmatrix}
                0 & 0 \\
                0 & 1 \\
            \end{pmatrix}$
        }& \multirow{2}{*}{
            $\begin{pmatrix}
                1 & 0 \\
                0 & 0 \\
            \end{pmatrix}$
        }\\
        & $y_{13}, y_{14} \in \g_+$ & & 0 & & \\
        \midrule
        \multirow{2}{*}{$x_2$ (hard)} & $y_{21}, y_{22} \in \g_\circ$ & \multirow{2}{*}{
            $\begin{pmatrix}
                1 & 0 \\
                1 & 1 \\
            \end{pmatrix}$
        } & -0.5 & \multirow{2}{*}{
            $\begin{pmatrix}
                1 & 0 \\
                0 & 0 \\
            \end{pmatrix}$
        } & \multirow{2}{*}{
            $\begin{pmatrix}
                0 & 0 \\
                1 & 0 \\
            \end{pmatrix}$
        } & \multirow{2}{*}{
            $\begin{pmatrix}
                0 & 0 \\
                1 & 0 \\
            \end{pmatrix}$
        }
        \\
        & $y_{23}, y_{24} \in \g_+$ &  & 0.5 & &  \\
        \bottomrule
    \end{tabular}
    \begin{tablenotes}
        \footnotesize
        \item[$\dagger$] CGPO case 1: for $x_2$ (hard), $y_{21} \in \g_\circ$ is accepted though $\g_+$ has greater expected return $\bar{r}(x_2, \g_+)$.
        \item[$\ddagger$] CGPO case 2: $y_{14}$ and $y_{23}$ are accepted, exceeding the density budget of $q_+ = 0.5$.
        \item[$\sharp$] Ours: $\textsc{opt}_\mathrm{IuB}$ accepts $y_{11}$ and $y_{23}$ so that margin is maximized ($r_\Delta(x_2, y_{23}) = 0.5$), and density cap $q_+$ holds.
    \end{tablenotes}
    \end{threeparttable}
\end{table}

{\bf Resulted update (IBPO).} Recap our update was defined as 
$\theta' = \theta - \alpha \nabla_\theta {\E}_{x\sim\mu_\xmat} \left[ \KL(  \textsc{sg}\{\hat{\pi}^\star_{\xmat, \ymat_\theta}\} \Vert \pi_\theta)[x]\right] $ 
subject to 
$\hat{\pi}^\star_{\xmat, \ymat_\theta} \in \argmax_{\pi\in \Pi\cap \hat{\Phi}_+(\xmat, \ymat_\theta)} \hat{\J}(\pi; \xmat, \ymat_\theta)$,
in Section~\ref{sec:design}. 
And we've now defined the RL objective $\J$ and the margin reward function $r_\Delta$. We are now ready to put everything together:
\begin{equation}
\begin{aligned}
\hat{\pi}^\star_{\xmat, \ymat_\theta} \in &\textstyle\argmax_{\pi} \textstyle\hat{\J}_\Delta(\pi; \xmat, \ymat_\theta) := \frac{1}{nm}\sum_i^n\sum_j^m [\pi(y_{ij}|x_{i})r_\Delta(x_i, y_{ij})] \\
&\underbrace{\quad\quad\quad\quad\quad \text{s.t.}~\pi \in \Pi \cap \hat{\Phi}_+(\xmat, \ymat_\theta) \text{~and~}
\textstyle\sum_y[\pi(y|x)\ind_{ \{ y\in\Xi_x \} }]\geq 1~\text{for all $x\in\xmat$}}_{\text{shorthanded as $\textsc{opt}_\mathrm{IuB}$ (inference under budget)}} \label{eq:final}
\end{aligned}
\end{equation}

In addition, as CGPO is a generalization of BoN (with tie breaking randomly), $\hat{\pi}^\star$ will be a pure strategy instead of stochastic one, meaning at most one $y$ will be accepted for each $x$, for subsequent projection (SFT).

{\bf Intuition.} Behind this update our intuition can be interpreted as: for a mini-batch $(\xmat, \ymat_\theta)$, $\hat{\pi}^\star$ aims to maximize the expected margin, while staying within the density constraints specified feasible region $\hat{\Phi}_+$; and in addition, only assign positive weight to responses $y \in \Xi_x$, meaning $y$ is both correct and within KL range.

Or verbally, if a prompt $x$ is difficult, the margin of extended responses $r_\Delta(x, y \in \g_+) = \bar{r}_\pi(x, \g_+) - \bar{r}_\pi(x, \g_\circ)$ is likely to be large for $y\in\g_+$. Therefore $\hat{\pi}^\star$ will more likely to assign positive weight to an extended response $y \in\g_+$ so that the objective receive more margin reward. On the other hand, if a query $x$ is simple, $r_\Delta(x, y \in \g_+)$ is likely to be small. Hence $\hat{\pi}^\star$ will possibly assign positive weight to regular responses $y\in\g_\circ$, so that one could save some density budget for harder queries. A concrete example is given in Table~\ref{tab:iub-example}.

{\bf Reward models.} 
Note that the original CGPO implementation has reward models for BoN ranking. We however intentionally excludes reward models, in our $\textsc{opt}_\mathrm{IuB}$ formulation as shown in Eq.~\eqref{eq:final}, to highlight our methodological contributions, by decoupling reward modeling efforts. 
Nonetheless, using reward models remains possible. 
To avoid introducing additional notation, we elaborate verbally: $\textsc{opt}_\mathrm{IuB}$ essentially selects either group $\g_+$ or $\g_\circ$ for a query $x$. 
Within a group $\g$, all responses receive the same reward $r_\Delta(x, \g)$, leaving it possible to further rank responses within each group using reward models.

{\bf Implementation \& solvers.} A pseudo-code of our IBPO with  $\textsc{opt}_\mathrm{IuB}$ is listed in Algorithm~\ref{algo:cgpo}. The essential change is to replace the constrained reward ranking $\textsc{opt}_\textsc{cgpo}$ with a general optimization problem, in our case the margin maximization under budget denoted as $\textsc{opt}_\mathrm{IuB}$. The $\textsc{opt}_\mathrm{IuB}$ problem is a (integer) linear programming problem that could be solved by off-the-shelf solvers, such as CPLEX~\citep{cplex2009v12}, Gurobi~\citep{gurobi} or SciPy~\citep{2020SciPy-NMeth}, which is our choice.

\begin{algorithm}[!t]
\caption{Inference Budget-Constrained Policy Optimization (IBPO)}
\begin{algorithmic}[1]
\REQUIRE 
prompt set $\mathbb{D}$, 
batch size $n$, 
number of responses $m$,  
init policy $\pi_0 = \pi_\mathrm{ref}$, 
num of iters $T$, 
budgets $q_+$
\FOR{$t = 1, \ldots, T$}
    \STATE {\bf prompt sampling} and {\bf response generation}: $\xmat^n \sim \mu_\mathbb{D}$ and $\ymat^{n\times m}_\theta \sim \pi_{\theta_t}(\xmat)$
    \STATE {\bf evaluate correctness} and {\bf empirical KL}: $\rmat^{n\times m}_\mathrm{match} = r_\mathrm{match}(\xmat, \ymat)$ and $\hat{\mathbf{KL}}^{n\times m} = \hat{\KL}(\ymat; \xmat, \pi_\mathrm{ref})$
    {\color{gray}
    \IF{CGPO (i.e. w/o IuB)}
        \STATE {\bf solve Best-of-$N$}: $\hat{\pi}^\star_{\xmat, \ymat_\theta} \in \textsc{opt}_\mathrm{cgpo}(\xmat, \ymat_\theta, \rmat_\mathrm{match}, \hat{\mathbf{KL}})$ as defined in Eq.~\eqref{eq:cgpo}
    \ENDIF
    }
    {\color{RoyalBlue}
    \IF{IBPO (i.e. w/ IuB)}
        \STATE {\bf margin maximization}: $\hat{\pi}^\star_{\xmat, \ymat_\theta} \in  \textsc{opt}_\textsc{iub}(\xmat, \ymat_\theta, \rmat_\mathrm{match}, \hat{\mathbf{KL}}, q_+)$ as defined in Eq.~\eqref{eq:final}, using solver 
    \ENDIF
    }
    \STATE {\bf gradient update}: with
    $-\sum_i\sum_{j}
    {\color{RoyalBlue} \hat{\pi}^\star_{\xmat, \ymat_\theta}(y_{ij} | x_i)}
    \frac{\partial}{\partial \theta}
    \log \pi_\theta(y_{ij} | x_i) \big|_{\theta = \theta_t} $
\ENDFOR
\end{algorithmic}\label{algo:cgpo}
\end{algorithm}

%% file: 4.construction.tex
\protected\def\myfavoriteat{%
  {\fontfamily{ptm}\selectfont @}%
}
\begingroup\lccode`~=`@ \lowercase{\endgroup\let~}\myfavoriteat
\catcode`@=\active

\input{tables/prompt}

\section{Acronyms, Na\"ive Construction \texorpdfstring{$\g_+$}{g+} \& Training Pipelines}\label{sec:sv}

Yet we work on abstract groups $\g_\circ$ and $\g_+$. In this section, we present the details of our constructions of extended length responses, i.e. the extended group $\g_+$. However developing long reasoning models is beyond the scope of this work, as our focus is on the constrained optimization of LLMs. Our constructions are for demonstrative purpose only. 
Due to the intricate details involved in prompts, datasets, and training pipelines, this section may appear somewhat dense. 
To make it more approachable, we have structured our writing in a way that readers can, if they wish, focus on the broader ideas without delving deeply into the specifics of constructions.
A TL;DR version of this section is provided below.


{\bf TL;DR.} 
We construct two types of illustrative extended responses: Sequential Voting (SV) and Adaptive Sequential Voting (ASV). Figure~\ref{fig:prompts} visually explains how these constructions are implemented. The goal of the SV is to establish a baseline that generates only responses in \(\g_+ \), thereby serving as an uni-modal comparator. SV scales roughly as well as vanilla majority voting (MV), aka self-consistency~\citep{wang2022self}.
In contrast, ASV outputs a mixture of responses of \( y \in \g_\circ \) and \( y \in \g_+ \). This allows the model to adaptively decide which type of response to produce based on the query. The goal of ASV is to further enable IBPO optimization, as IBPO implicitly assumes the model generates both regular and extended-length responses. 
In Section~\ref{sec:exp}, we show that ASV optimized by IBPO, achieves better allocation of the inference budget.

\subsection{Construction of Sequential Vote}\label{sec:constructions}

{\bf Acronyms.} 
For clarity, we explicitly define key terms to hopefully resolve any potential ambiguities. 
{\it Response}: A response refers to a sequence generated until a terminal token is encountered. For precision, we sometimes refer to these as voting responses or SCoT responses, as illustrated in Figure~\ref{fig:prompts}, after introducing our sequential voting baselines.
{\it Trial}: A trial denotes a solution instance, which is demarcated by the special tokens [TRIAL] … [/TRIAL], as shown in the voting response example in Figure~\ref{fig:prompts}. While a voting response contains multiple trials, a SCoT response or a non-voting response contains exactly one trial, as also depicted in Figure~\ref{fig:prompts}.

For the description of training and testing details, 
we use LLaMA and LLaMA-b to denote the instruction-tuned and base versions of the LLaMA 3.1 8B models~\citep{dubey2024llama}, respectively. 
MATH refers specifically to the training split of the Hendrycks MATH dataset~\citep{hendrycks2021measuring}, while the 500-sample subset of the testing split is referred to as MATH500~\citep{lightman2023let}. 
SDPO stands for step-DPO~\citep{lai2024step} dataset, a curated step-annotated dataset from which we retain only the prompts and positive responses (ground truth solutions), excluding any step signals (see Appendix~\ref{sec:data} for details). It is important to note that while we leverage their curated dataset, the original SDPO method is not relevant to this work. 
The SDPO dataset was chosen because its ground truth responses follow the SCoT format of LLaMA responses, making it convenient to run supervised fine-tuning (SFT) mixed with LLaMA samples.

\input{tables/data_table}

\input{tables/training_table}

{\bf Construction details.} To be more specific, the na\"ive sequential voting baselines, as the ``expensive'' group $\g_+$, have increased inference costs by simply sequentially output multiple trials and find the sequential majority vote until stop condition met. 
In particular, the early-stopping sequential voting (SV) baseline is created to show that such na\"ive baseline could achieve performance gain, 
on par with vanilla majority voting (MV). The adaptive sequential voting (ASV) allows model to output both SCoT response $y_\circ$ and sequential voting response  $y_+$, allowing us to further conduct the budget controll experiments in Section~\ref{sec:exp}.
\begin{itemize}
    \item Early Stopping Sequential Vote (SV): For a single response, model are allowed to output at most 8 trials, and conclude with majority of trials. In addition to the terminal condition of maximum 8 trials, model will early stop if an answer appears 3 times and this answer will be considered as the majority answer. 
    \item Adaptive Sequential Vote (ASV): Model is allowed to choose either vote (case 1, i.e. $y_+$)  or not (case 2, i.e. $y_\circ$). This baseline is created to further allow model-driven resource allocation. We later, in Section~\ref{sec:exp}, show that one could optimize the ability of resource allocation with our IBPO.
\end{itemize}

{\bf Dataset.} 
We define our construction of dataset as a product of a problem set $\mathbb{Q}$, format template of question $\mathcal{T}_q$ and answer $\mathcal{T}_a$, and a response set $\mathbb{A}$, subjected to some filtration $\mathcal{F}$. 
Formally, a dataset $\D$ is defined as: $\mathbb{D} := (\mathcal{F} \circ \mathcal{T}_q \circ \mathcal{T}_a)(\mathbb{Q} \times \mathbb{A})$, 
where $\mathcal{F}$ removes undesired question-answer pairs, such as incorrect responses when $\mathbb{A}$ are model generated; 
The templates $\mathcal{T}_q$ and $\mathcal{T}_a$ collectively transform each question-answer pair into a specific text format, as shown in Figure~\ref{fig:prompts}; 
Slightly abusing the notation, the Cartesian product $\mathbb{Q} \times \mathbb{A}$ is used to pair each response with its corresponding question, defined as: $\mathbb{Q}\times \mathbb{A} := {\{(q_i, a_{ij}): \forall i, j \}}$.

Specifically, we summarize the datasets used in subsequent sections in Table~\ref{tab:datasets}. For instance, for SV training, we construct $\D_\textsc{sv}$ with question set $\mathbb{Q}_\textsc{math}$ from~\citet{hendrycks2021measuring} and LLaMA generated responses $\mathbb{A}^\textsc{sample}_\textsc{math}$ using the SV templates defined in Figure~\ref{fig:prompts}, subjected to data selection described in Appendix~\ref{sec:data}.

\begin{figure}[!t]
    \centering
    \begin{minipage}{0.48\textwidth}
        \centering
        \input{tables/benchmark_table}
    \end{minipage}
    \hfill
    \begin{minipage}{0.48\textwidth}
        \centering
        \begin{subfigure}{\textwidth}
            \includegraphics[width=0.885\textwidth]{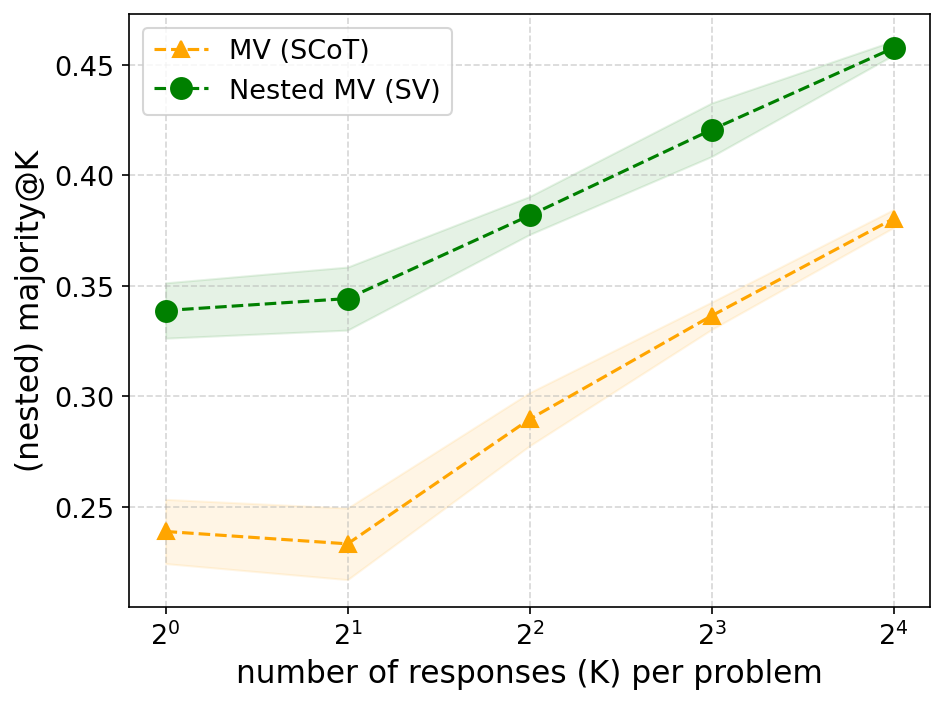}
            \caption{(Nested) MV measured with number of responses}
            \label{fig:nestd-mv-per-response}
        \end{subfigure}
        \hfill
        \begin{subfigure}{\textwidth}
            \vspace{1em}
            \includegraphics[width=0.92\textwidth]{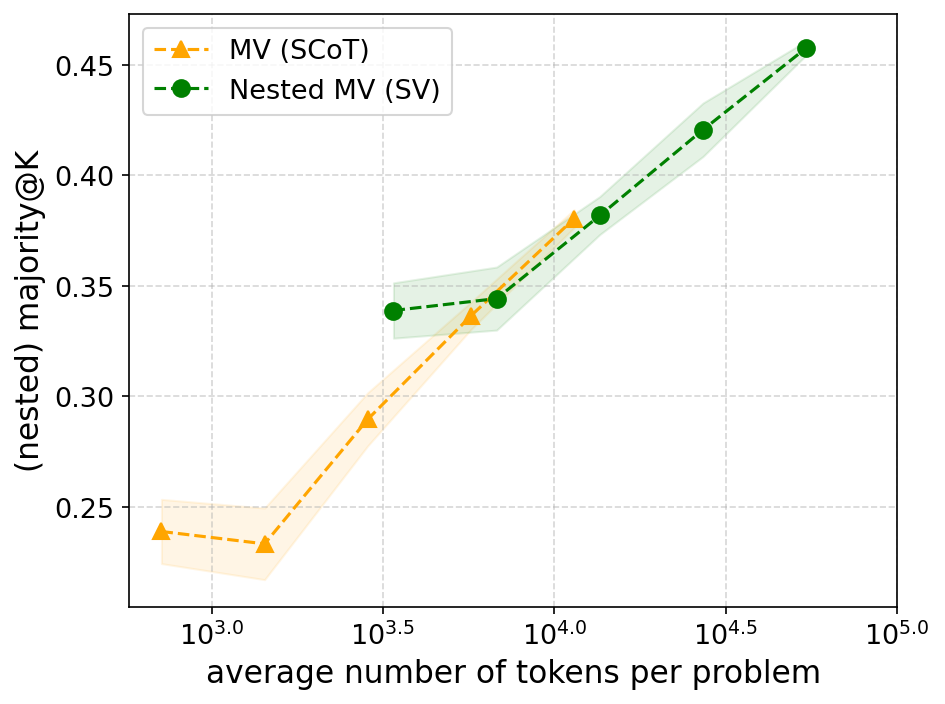}
            \caption{(Nested) MV measured with number of tokens}
            \label{fig:nestd-mv-per-token}
        \end{subfigure}
        \caption{SV tested on MATH500. MV stands for vanilla majority voting, aka self-consistency~\citep{wang2022self}, with SCoT responses. Nested MV are majority voting with our (early-stopping) SV responses. It is ``nested'' as each SV response is already an voting, as shown in Figure~\ref{fig:prompts}. The SV method have ``clear'' gain when performance is measured with number of responses. When measured with number of tokens, SV aligns the performance-cost efficiency of vanilla MV. This is another indicator that one should worry about token efficiency when measure reasoning performance.}
        \label{fig:mv-figures}
    \end{minipage}%
\end{figure}

{\bf Training pipelines.} We summarize our training pipelines for our toy experiments in Section~\ref{sec:exp-sv} and our IuB experiments in Section~\ref{sec:exp}. For instance, experiment 2.2 (Section~\ref{sec:exp}) in Table~\ref{tab:pipelines} is the summary of our IBPO training, using dataset $\mathbb{D}_\textsc{rl}$ as constructed in Table~\ref{tab:datasets} and initialized from ASV models from experiment 2.2. Further details of each training can be found in Appendix~\ref{sec:data} and~\ref{sec:hyperparams}.

\subsection{Performance of Nai\"ve Sequential Vote}\label{sec:exp-sv}

We start by evaluating the performance of SV compared to vanilla MV, following the setup of Exp. 1.1 in Table~\ref{tab:pipelines}. 
This toy experiment is designed to demonstrate: 
(i) SV scales approximately as well as MV, thereby qualifying it as an example of \(\g_+\); and
(ii) measuring performance based on the number of responses is inadequate, therefore we later measure performance relative to the number of tokens/trials in Section~\ref{sec:exp}.

\textbf{Metrics.} 
The metrics we use are pass@$k$ and majority@$k$, both of which are widely used in the literature \citep{hendrycks2021measuring, wang2022self}. 
We occasionally refer to pass rate as pass@$1$. In both metrics, $k$ specifically denotes the number of responses, regardless of the number of trials per response. 
Since our voting methods may involve multiple trials, we may use the average number of trials (as illustrated in Figure~\ref{fig:pass-rate}) on the x-axis for cost-aware comparisons. 
In addition, we evaluate “performance-cost” efficiency by comparing each method’s scaling efficiency to that of MV, following the comparison in~\citet{snell2024scaling}.

As shown in Figure~\ref{fig:nestd-mv-per-response}, SV exhibits significant improvements—nearly 10\%—in terms of majority@$k$ when performance is measured by the number of responses. 
However, this improvement is misleading since each SV response effectively consists of multiple SCoT responses. 
Therefore, measuring performance with taking length into consideration provides a more reasonable assessment. 
As demonstrated in Figure~\ref{fig:nestd-mv-per-token}, SV’s scaling performance aligns with that of vanilla MV, supporting its role as a suitable construction example for \(\g_+\).

%% file: tables/prompt.tex
\begin{figure}[!t]
\begin{minipage}{0.5\textwidth}
\begin{tcolorbox}[title=\textbf{SCoT Prompt + Cond.}]
\small
Solve the following math problem efficiently and clearly: 

- For simple problems (2 steps or fewer): 

Provide a concise solution with minimal explanation. 

- For complex problems (3 steps or more): 

Use this step-by-step format: 

\#\# Step 1: [Concise description] 

[Brief explanation and calculations] 

\#\# Step 2: [Concise description] 

[Brief explanation and calculations] 

... 

Regardless of the approach, always conclude with: 

Therefore, the final answer is: \$\textbackslash\textbackslash boxed\{answer\}\$. I hope it is correct. 

Where [answer] is just the final number or expression that solves the problem. 

{\color{RoyalBlue} [Condition Description]}

Problem: \{\{ problem \}\}
\end{tcolorbox}

\begin{tcolorbox}[title=\textbf{SCoT Response}]

\small 

\#\# Step 1: {\color{RoyalBlue} ... steps omitted}

The final answer is: $\boxed{\textsc{A1}}$. I hope it is correct.

\end{tcolorbox}

\begin{tcolorbox}[title=\textbf{SV Cond.}]

\small

You are asked to give at most eight diverse solutions in different way, without referencing to the previous trials.

Each trial should be contained in a separate [TRIAL] trial solution [/TRIAL] block.

If a solution occurs three times, it is considered as a consensus and will be used as the final answer.

If there is no consensus, use the solution from the most plausible trial.
\end{tcolorbox}
\end{minipage}
\hfill
\begin{minipage}{0.5\textwidth}
\begin{tcolorbox}[title=\textbf{ASV Cond.}]
\small

For medium and hard level problems, you are asked to give at most eight diverse solutions in different way, without referencing to the previous trials.

Each trial should be contained in a separate [TRIAL] trial solution [/TRIAL] block.

If a solution occurs three times, it is considered as a consensus and will be used as the final answer.

If there is no consensus, use the solution from the most plausible trial.

For easy level problems, you are allowed only one attempt, which will be considered your final answer.

\end{tcolorbox}
\begin{tcolorbox}[title=\textbf{Voting Response (SV, ASV Case 1)}]

\small 

[TRIAL] \#\# Step 1: {\color{RoyalBlue} ... steps omitted}

The final answer is: $\boxed{\textsc{A1}}$. [/TRIAL] \\

[TRIAL] \#\# Step 1: {\color{RoyalBlue} ... steps omitted}

The final answer is: $\boxed{\textsc{A2}}$. [/TRIAL] \\

[TRIAL] \#\# Step 1: {\color{RoyalBlue} ... steps omitted}

The final answer is: $\boxed{\textsc{A1}}$. [/TRIAL] \\

[TRIAL] \#\# Step 1: {\color{RoyalBlue} ... steps omitted}

The final answer is: $\boxed{\textsc{A1}}$. [/TRIAL] 

The answer $\boxed{\textsc{A1}}$ has occurred three times, and is considered as a consensus.

The final answer is $\boxed{\textsc{A1}}$. I hope it is correct.

\end{tcolorbox}
\begin{tcolorbox}[title=\textbf{Non-Voting Response (ASV Case 2)}]

\small 

[TRIAL] \#\# Step 1: {\color{RoyalBlue} ... steps omitted}

The final answer is: $\boxed{\textsc{A1}}$. [/TRIAL] 

Terminated due to difficulty level.

\end{tcolorbox}

\end{minipage}
\caption{Prompt templates. 
For $\g_\circ$, we use the SCoT prompt without including an additional [condition description] and generate a standard SCoT response. For $\g_+$, we simply insert the corresponding condition into the [condition description] placeholder to create a new prompt. In the case of SV, the SV condition is integrated into the prompt, and the model is asked to perform repeated trials to reach a consensus. The ASV prompt instructs the model to output either a voting response (case 1) or a non-voting response (case 2), with the decision made by the model itself.}
\label{fig:prompts}
\end{figure}

%% file: tables/data_table.tex
\begin{table}[!t]
    \centering
    \begin{minipage}{0.5\textwidth}
        \centering
        \begin{threeparttable}
            \caption{Summary of constructed datasets for different experimental purposes. $\mathbb{D}_\textsc{sv}$, $\mathbb{D}_\textsc{essv}$ and $\mathbb{D}_\textsc{asv}$ uses the same set of prompts but different prompt/response templates. $\mathbb{Q}_\textsc{sdpo}$ and $\mathbb{A}^\textsc{golden}_\textsc{sdpo}$ are from~\citet{lai2024step} with details deferred to Appendix~\ref{sec:data}. ASV1 and ASV2 correspond to ASV case 1 and 2 (see Section~\ref{sec:constructions}) respectively.}
            \label{tab:datasets}
            \begin{tabular}{l|l|l|l|l|l}
                \toprule
                Set & $\mathcal{T}_q$ & $\mathcal{T}_a$ & $\mathcal{F}$ & $\mathbb{Q}$ & $\mathbb{A}$ \\
                \midrule
                $\mathbb{D}_\textsc{sv}$  & SV & SV & \multirow{2}{*}{~\ref{sec:data}~}  & \multirow{2}{*}{$\mathbb{Q}_\textsc{math}$} & \multirow{2}{*}{$\mathbb{A}_\textsc{math}^\textsc{sample}$} \\
                $\mathbb{D}_\textsc{asv1}$ & ASV & ASV1 &   &  &  \\
                \midrule
                $\mathbb{D}_\textsc{asv2}$ & ASV & ASV2 & - & \multirow{2}{*}{$\mathbb{Q}_\textsc{sdpo}$} & \multirow{2}{*}{$\mathbb{A}_\textsc{sdpo}^\textsc{golden}$} \\
                $\mathbb{D}_\textsc{scot}$ & SCoT & SCoT & - &  & \\
                \midrule
                $\mathbb{D}_\textsc{rl}$ & ASV & - & - &  
                $\mathbb{Q}_\textsc{sdpo}$
                  & $\emptyset$ \\
                \bottomrule
            \end{tabular}
        \end{threeparttable}
    \end{minipage}%
    \hfill
    \begin{minipage}{0.48\textwidth}
        \centering
        \begin{minipage}{\linewidth}
        \footnotesize
        \begin{threeparttable}
            \captionsetup{skip=1em}
            \caption{Stopping conditions.}
            \label{tab:rules}
            \begin{tabular}{l|p{0.8\textwidth}}
                \toprule
                 & stopping conditions \\ 
                \midrule
                SV & (i) max 8 trials; (ii) if an answer occurs 3 times. \\ 
                \midrule
                ASV & voting (case 1): (i) max 8 trials; (ii) if an answer occurs 3 times; non-voting (case 2): exact 1 trial. \\
                \bottomrule
            \end{tabular}
        \end{threeparttable}
        \end{minipage}\\
        \begin{minipage}{\linewidth}
        \footnotesize
        \begin{threeparttable}
            \captionsetup{skip=1em} 
            \caption{Prompt and response sources.}
            \label{tab:dummy}
            \vspace{2em}
            \begin{tabular}{l|l|c}
                \toprule
                Set & Query-Response & Source \\
                \midrule
                \multirow{2}{*}{Training} 
                & $\mathbb{Q}_\textsc{math}$ \& $\mathbb{A}_\textsc{math}^\textsc{sample}$ & MATH \& LLaMA samples \\
                & $\mathbb{Q}_\textsc{sdpo}$ \& $\mathbb{A}_\textsc{sdpo}^\textsc{golden}$ & {\citep{lai2024step}} \\
                \midrule
                Testing & $\mathbb{Q}_\textsc{math500}$ & MATH500 \\
                \bottomrule
            \end{tabular}
        \end{threeparttable}
        \end{minipage}
    \end{minipage}
\end{table}

%% file: tables/training_table.tex
\begin{table}[!t]
    \centering
    \resizebox{\textwidth}{!}{
    \begin{threeparttable}
        \caption{Training pipelines. For Sec.~\ref{sec:exp-sv}, we aim to create a demonstration experiment showing that with the same model (of roughly same math knowledge) SV can achieve reasonable performance-cost efficiency on par with MV. Sec.~\ref{sec:exp} further show we can optimize performance-cost efficiency through our IuB generalization of CGPO, where $\alpha$ in row 2.1 is a coefficient of $\mathbb{D}_\textsc{asv2}$.}
        \begin{tabular}{l|l|l|l|l|l}
            \toprule
            Exp. & Sec. & Type & Init. Ckpt. & Dataset & Purpose  \\
            \midrule
            1.1 & Sec.~\ref{sec:exp-sv} & SFT & LLaMA-b & \multirow{2}{*}{$\mathbb{D}_\textsc{sv} \cup \mathbb{D}_\textsc{scot}$} & \multirow{2}{*}{\small Allow model follow both SV and SCoT prompt.}  \\
            1.2 & Sec.~\ref{sec:exp} & SFT & LLaMA & & \\
            \midrule
            2.1 & \multirow{2}{*}{Sec.~\ref{sec:exp}} & SFT & LLaMA & $\mathbb{D}_\textsc{asv1} \cup \alpha \mathbb{D}_\textsc{asv2}$ &  \begin{tabular}{@{}l@{}} \small Follow the ASV instruction to let model decide vote or not. \end{tabular}  \\       
            2.2 & & RL & Exp. 2.1 & $\mathbb{D}_\textsc{rl}$ &  \begin{tabular}{@{}l@{}} \small Optimize the capability of dynamic budget allocation. \end{tabular}  \\  
            \bottomrule
        \end{tabular}\label{tab:pipelines}
        
    \end{threeparttable}    }
    
\end{table}

%% file: tables/benchmark_table.tex
\footnotesize
\centering
\captionof{table}{Comparison of approaches on their improvements versus base models. $^\dagger$, $\ddagger$, $\dagger\dagger$ indicate results duplicated from~\citet{qu2024recursive, yan2024s, kumar2024training}, respectively. $^*$ indicates our methods/constructions and the improvements are relative to LLaMA model.}
    \centering
    \resizebox{\textwidth}{!}{
    \begin{tabular}{l|cc|c}
    \toprule
    approach & \begin{tabular}{@{}c@{}} pass@$1$  \end{tabular}  & improv. & \begin{tabular}[c]{@{}c@{}} turns/trials \\ per response \end{tabular} \\
    \midrule
    \multicolumn{4}{c}{SFT/Prompting-based} \\
    \midrule
    {\bf SV-SFT}$^*$ & & & \\ 
    \phantom{~~}LLaMA & 56.8 & 5.54 & 5.67x \\
    {\bf ASV-SFT-1}$^*$ & & & \\ 
    \phantom{~~}LLaMA & 55.6 & 4.43 & 5.74x \\
    \midrule
    {\bf SFT-RISE$^\dagger$} & & & \\
    \phantom{~~}setting 1 (table 1 of $\dagger$) & 5.5 & -0.3 & 5x \\
    \phantom{~~}setting 2 (table 1 of $\dagger$) & 5.0 & 0.0 & 5x \\
    \midrule
    {\bf SFT-SCoRe$^{\dagger\dagger}$} & & & \\
    \phantom{~~}setting 1 (table 1 of ${\dagger\dagger}$) & 54.2 & 1.8 & 2x \\
    \phantom{~~}setting 2 (table 1 of ${\dagger\dagger}$) & 55.0 & 0.0 & 2x \\
    \midrule
    {\bf RISE$^\dagger$} & & & \\
    \phantom{~~}LLaMA2 Base & 1.4 & -0.5 & 5x \\
    \phantom{~~}+ boosting & 5.5 & 0.0 & 5x \\
    \midrule
    {\bf S$^3$C$^\ddagger$} & & & \multirow{5}{*}{not studied}\\
    \phantom{~~}LLaMA3-8B  & 33.14 & 2.56 &  \\
    \phantom{~~}Mistral-7B & 25.48 & 1.44 &  \\
    \phantom{~~}{\scriptsize DeepSeek-Math-Base-7B} & 41.40 & 3.18 &  \\
    \phantom{~~}Qwen2-Math-7B & 51.76 & 0.44 &  \\
        \midrule
    {\bf Self-Refine$^\dagger$} & & & \\
    \phantom{~~}Base            & 1.9 & 0.0 & 3x \\
    \phantom{~~}GPT-3.5         & 36.5 & -3.2 & 3x \\
    \phantom{~~}Mistral-7B      & 7.1 & -0.4 & 3x \\
    \phantom{~~}Eurus-7B-SFT    & 9.0 & -3.3 & 3x \\
    \midrule
    {\bf STaR$^{\dagger\dagger}$} & & & \\
    \phantom{~~}setting 1 (table 1 of ${\dagger\dagger}$) & 54.0 & 0.4 & 2x \\
    \phantom{~~}setting 2 (table 1 of ${\dagger\dagger}$) & 41.2 & -14.2 & 2x \\
    \midrule
    \multicolumn{4}{c}{Online Iterative/RL} \\
    \midrule
    {\bf ASV-IuB-$q_+$}$^*$ & & & \\
    \phantom{~~}$q_+ = 25\%$ & 54.2 & 2.94 & 2.24x \\
    \phantom{~~}$q_+ = 50\%$ & 55.4 & 4.14 & 2.16x \\
    \phantom{~~}$q_+ = 75\%$ & 57.0 & 5.74 & 4.32x \\
    \midrule
    {\bf RISE$^\dagger$} & & & \\
    \phantom{~~}+ Iteration 1 & 9.7 & 3.4 & 5x \\
    \phantom{~~}+ Iteration 2 & 10.4 & 4.6 & 5x \\
    \midrule
    {\bf SCoRe$^{\dagger\dagger}$} & & & \\
    \phantom{~~}Gemini 1.5 Flash & 64.4 & 4.4 & 2x \\
    \phantom{~~}+ more turns {(fig. 8$^{\dagger\dagger}$)} & $\approx$66 & $\approx$6 & 5x-10x \\
    \bottomrule
    \end{tabular}}
\label{tab:performance_metrics}

%% file: 5.performance.tex
\section{Evaluation of IBPO w/ \texorpdfstring{$\textsc{OPT}_\mathrm{IuB}$}{opt\_IuB}}\label{sec:exp}

\subsection{Absolute Improvement (Table~\ref{tab:performance_metrics})}
 
{\bf ``Baselines''.} To put SV/ASV in comparison with other baselines in literature, we gather several baselines from the self-correction literature as essentially these methods increase inference length, though not extensively as~\citet{jaech2024openai}. The results in Table~\ref{tab:performance_metrics} are mainly gathered  from~\citet{qu2024recursive, kumar2024training}. The self-correction baselines often admit a multi-turn structure, similarly to the multi-trial construction of SV. We therefore include a column of inference cost measured by number of turns/trials for a rough comparison.
The SFT comparators we include are (reproduced) Self-Refine~\citep{madaan2024self}, STaR~\citep{zelikman2022star}, S$^3$C~\citep{yan2024s}, SFT-RI and SFT-SCoRe, where SFT-RI and SFT-SCoRe are SFT comparators implemented in Recursive Introspection (RI)~\citep{qu2024recursive} and SCoRe~\citep{kumar2024training} respectively.

{\bf ASV experiments.} SV-SFT follows our training pipeline of Exp 1.2 in Table~\ref{tab:pipelines} and is an voting only baseline. ASV-SFT-$\alpha$ follows our Exp 2.1 setup and is an adpative baseline, meaning model decide whether to vote or not for a problem. We report ASV-SFT-$1$ only as it is empirically the best adaptive baseline although still fall short in terms of efficiency. Our ASV-IuB-$q_+$ experiments, all initialized from ASV-SFT-$1$, are our adaptive models, optimized by IBPO with $\textsc{opt}_\mathrm{IuB}$ as lower level optimization, e.g. Algorithm~\ref{algo:cgpo}.

{\bf Observations.} 
We would like to first emphasize that the comparison in Table~\ref{tab:performance_metrics} is not intended to demonstrate that SV outperforms SFT-based self-correction or that our ASV-IuB surpasses RL-based self-corrections. These efforts are orthogonal, as our focus is on constrained optimization.
As observed, our SFT constructions—SV-SFT and ASV-SFT-1 achieve a clear improvement in pass@$1$ with high inference costs ($5+$ times the number of trials).
The ASV-IuB-$q_+$ formulation, particularly with $q_+ = \{50\%, 75\%\}$, shows significant improvement while reducing costs by $4.14\%$ at $2.16\times$ and $5.74\%$ at $4.32\times$. This performance is on par with SCoRe, a state-of-the-art RL-based self-correction method.
{Note that the performance of ASV-IuB-$q_+$ is reported using the best checkpoints. Results from the last checkpoint are shown in Figure~\ref{fig:pass-last}. Additionally, training curves are presented in Appendix~\ref{sec:further-discussion}, which shows consistent improvements.}
As a somewhat tangential yet potentially intriguing observation, it is evident that prompting-based and SFT-based methods struggle with both absolute improvement (Table~\ref{tab:performance_metrics}) and efficiency (Figure~\ref{fig:pass-rate}), supporting the conjecture that SFT alone does not enable self-correction capabilities~\citep{huang2023large, kumar2024training}.
This observation is also partially supported by concurrent work~\citep{deepseekai2025deepseekr1incentivizingreasoningcapability}, which suggests that such self-correction behavior emerges automatically during RL rather than manually created by prompting or SFT.

\begin{figure}[!t]
    \centering
    \begin{subfigure}[b]{0.485\textwidth}
        \includegraphics[width=\textwidth]{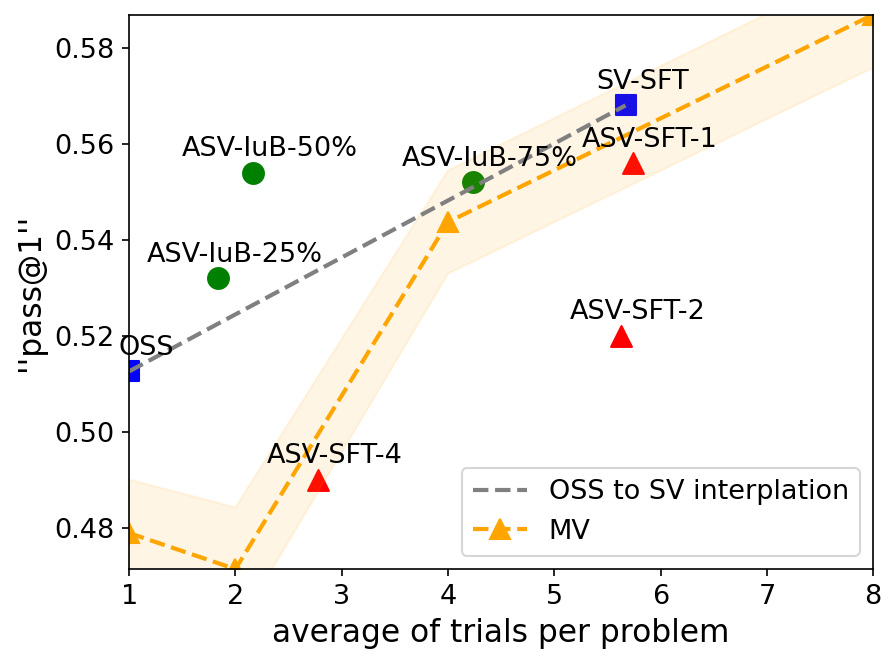}
        \caption{performance of last checkpoints}
        \label{fig:pass-last}
    \end{subfigure}
    \hfill
    \begin{subfigure}[b]{0.485\textwidth}
        \includegraphics[width=\textwidth]{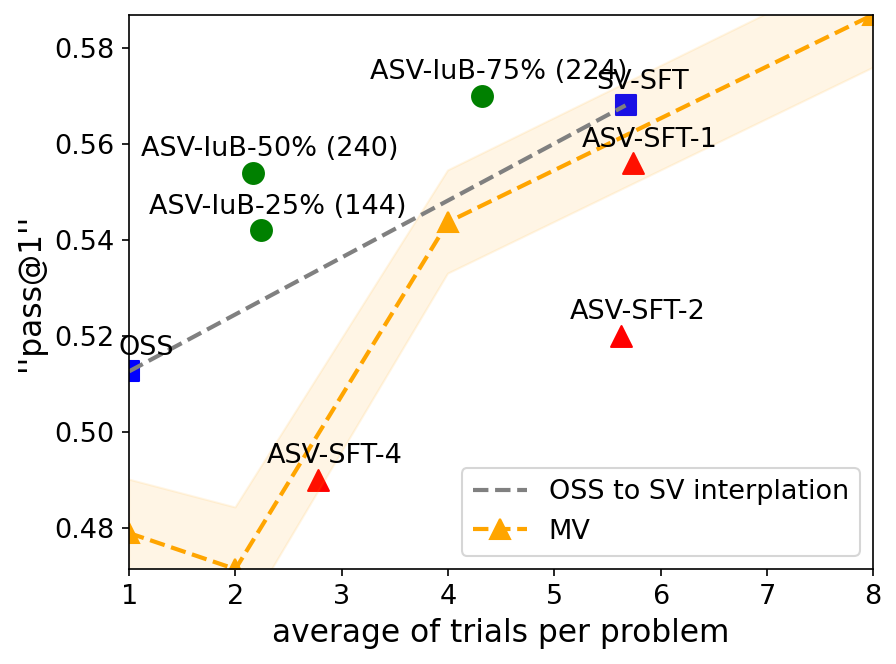}
        \caption{best checkpoints (with steps in parentheses)}
        \label{fig:pass-best}
    \end{subfigure}
    \caption{Comparison of pass@$1$ against the average number of trials (majority@$N$ for MV). OSS refers to LLaMA. The interpolation between OSS and SV-SFT (aligning with MV) serves as a hypothetical efficiency boundary. ASV-SFT shows lower efficiency relative to this boundary, whereas ASV-IuB consistently achieves higher efficiency.}
    \label{fig:pass-rate}
\end{figure}

\begin{figure}[!t]
    \centering
    \begin{subfigure}[b]{0.485\textwidth}
        \includegraphics[width=\textwidth]{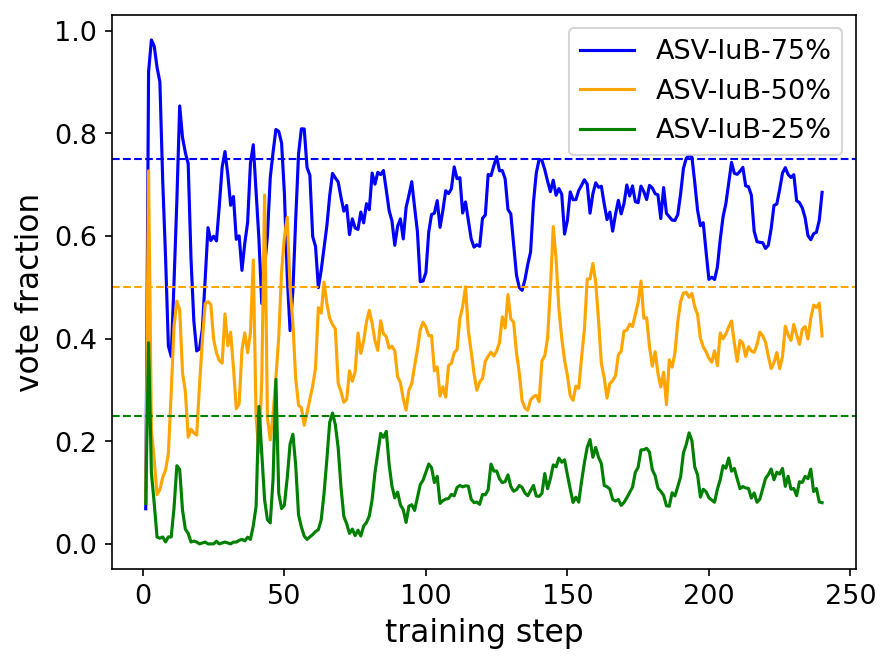}
        \caption{voting ratio of correct responses (training set)}
        \label{fig:budgets-train}
    \end{subfigure}
    \hfill
    \begin{subfigure}[b]{0.485\textwidth}
        \includegraphics[width=\textwidth]{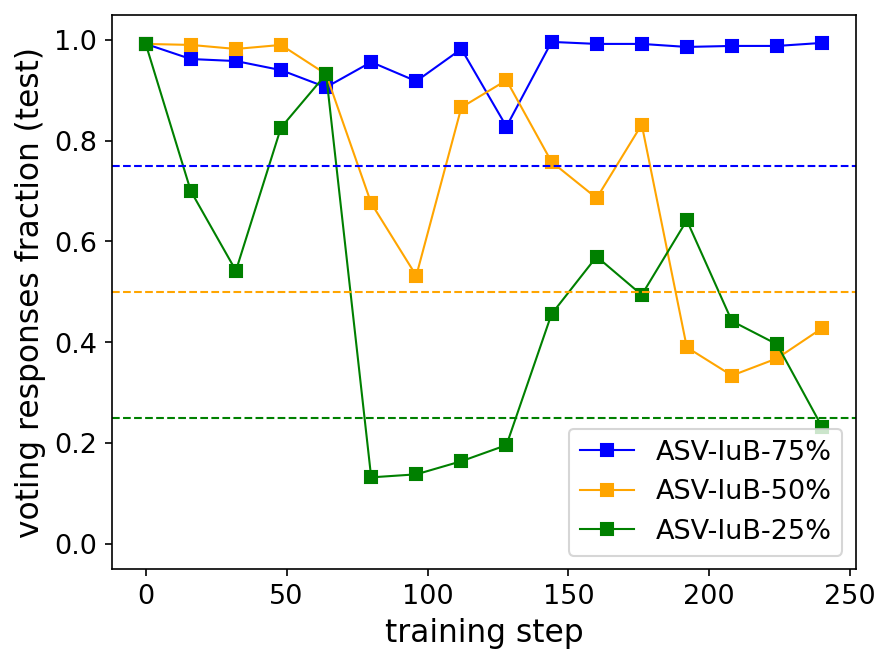}
        \caption{voting response ratio (testing set)}
        \label{fig:budgets-test}
    \end{subfigure}
    \caption{Voting response ratio versus training steps. Dashed line denotes the budget $q_+$. On the training set, IuB formulation follows the budget constraints almost exactly. Due to distribution shift, the constraint on testing set is not entirely exact, but still it is noticeable that the voting ratio follows the order of $75\% \succ 50\% \succ 25\% $.}
    \label{fig:budgets}
\end{figure}

\begin{figure}[!t]
    \centering
    \begin{subfigure}[b]{0.485\textwidth}
        \includegraphics[width=\textwidth]{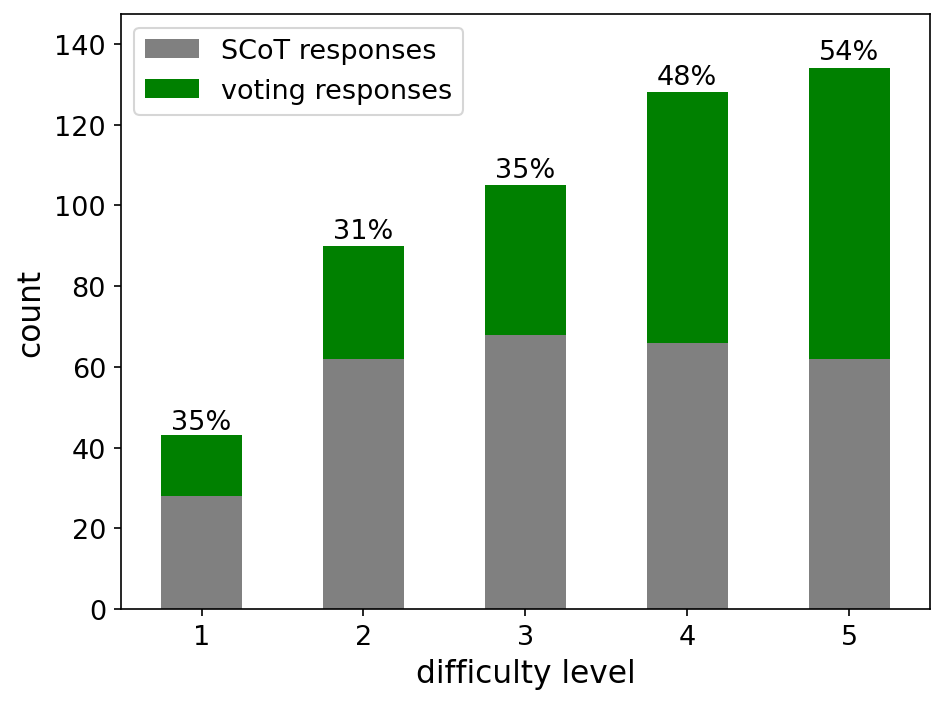}
        \caption{voting response ratio grouped by difficulty ($q_+ = 50\%$)}
        \label{fig:difficulty-50}
    \end{subfigure}
    \hfill
    \begin{subfigure}[b]{0.485\textwidth}
        \includegraphics[width=\textwidth]{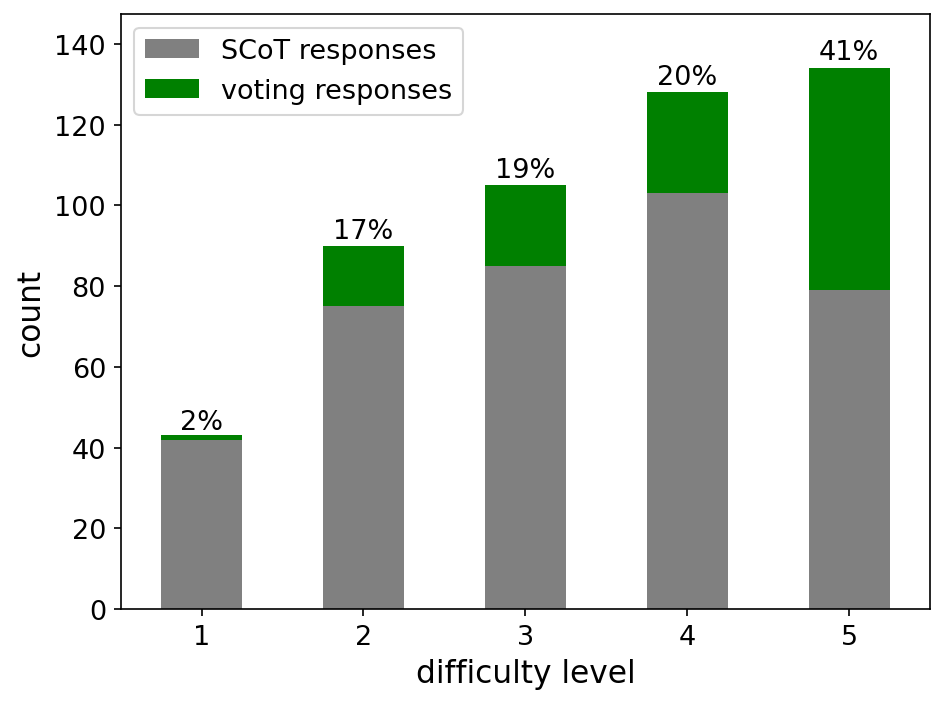}
        \caption{voting response ratio grouped by difficulty ($q_+ = 25\%$)}
        \label{fig:difficulty-25}
    \end{subfigure}
    \caption{The IuB formulation enables the model to dynamically allocate voting budget to harder problems.}
    \label{fig:difficulty}
\end{figure}

\subsection{Efficiency, Constraint Following \& Budget Allocation}

Table~\ref{tab:performance_metrics} demonstrates that our ASV-IuB-$q+$ models can achieve performance comparable to RL-based self-correction models simply through inference allocation management. In this subsection, we elaborate on our discussion of (i) performance-budget efficiency, (ii) constraint satisfaction, and (iii) inference budget allocation.

{\bf Performance-budget efficiency.} 
In Figure~\ref{fig:pass-rate}, we visually assess the performance-budget efficiency, compared to a hypothetical efficiency boundary. 
This boundary is an interpolation between OSS LLaMA model and SV-SFT. It is reasonable to see it as a hypothetical boundary for two reasons: (i) OSS and SV-SFT are two extremes of ASV-IuB-$q_+$, corresponding to the cases of $q_+ = 0$ and $q_+ = 1$ respectively; and (ii) this interpolation achieves an increase comparable to MV, if not slightly better.
The SFT version of ASV is generally much worse than the boundary, as SFT alone does not optimize resource allocation as a mathematical objective.
For ASV-IuB-$q_+$ optimized by IBPO, we report both the last and best checkpoint results in Figure~\ref{fig:pass-last} and Figure~\ref{fig:pass-best}, respectively.
Our formulation achieves, in general, better performance-budget efficiency, except $q_+ = 75\%$ in Figure~\ref{fig:pass-last}. We will extend our discussion on this unsuccessful case in subsequent paragraphs.

{\bf Constraint satisfaction.} 
We then evaluate how effectively the constraints are enforced. In Figure~\ref{fig:budgets-train}, 
the budget constraints are successfully maintained during training for $q_+ = \{25\%, 50\%, 75\%\}$. 
Due to distribution shifts, exact adherence to these constraints is not expected on the test set. 
Nevertheless, Figure~\ref{fig:budgets-test} demonstrates that constraints are still upheld at the end of training for $q_+ = \{25\%, 50\%\}$, 
and the ratio of voting responses follows the order $75\% \succ 50\% \succ 25\%$. 
(Figure~\ref{fig:budgets-train} illustrates that our model meets the budget constraints for the set of correct responses. The constraints also hold for all responses, see Appendix~\ref{sec:further-discussion}.)

{\bf Difficulty-aware allocation.} 
While we have demonstrated improved efficiency and adherence to constraints, we would like to validate the underlying intuition of our design: 
that more challenging problems may require longer reasoning steps, whereas simpler problems can be resolved with just SCoT responses. 
Ideally, the model should allocate more voting responses to problems with higher difficulty levels.
To this end, we use the difficulty levels from the metadata of Hendrycks MATH and plot the ratio of voting responses for each difficulty level. 
Figure~\ref{fig:difficulty} illustrates that for both $q_+ = \{25\%, 50\%\}$, more challenging problems, such as those at levels 4 and 5, receive a higher allocation of budgets.
This allocation pattern is particularly evident for the case of $q_+ = 25\%$, where only $2\%$ of level 1 problems receive voting responses.

{\bf Discussion on the unsuccessful case.} 
There is one observed unsuccessful case in Figure~\ref{fig:pass-last}: the last checkpoint of ASV-IuB-$75\%$, 
which falls on the hypothetical boundary rather than above it. 
This outcome is arguably expected, as observed in Figure~\ref{fig:budgets-test}, where ASV-IuB-$75\%$ outputs almost exclusively voting responses at the end of training. 
As a result, this model is not adaptively allocating resources, and thus no improvement in efficiency is anticipated. 
This unsuccessful case is hence caused by the distribution shift between the training set and the testing set.
It is possible that scaling the training set—given that our training set $\D_\textsc{rl}$ contains approximately $10,000$ prompts, 
which is relatively small—will make the testing set more likely to be in-distribution and thereby alleviate the distribution shift issue.

%% file: 6.conclusion.tex
\section{Conclusion \& Discussions}\label{sec:con}

We derived a constrained policy optimization framework, IBPO, from an optimization perspective, resulting in a simple weighted SFT update that resembles successful iterative SFT algorithms such as RFT and RAFT. In each iteration, the optimal weight is determined by solving an (integer) linear program. The practical implementation of IBPO is build on top of CGPO, and is evaluated on a math reasoning task with inference budget constraints. Empirical evaluations demonstrate that our framework enables the model to adhere to constraints and dynamically allocate the inference budget.

{\bf Batch optimization.} Since we solve an optimization problem per iteration (i.e. per mini-batch), limited computational resources can result in smaller sample sizes for the inner optimization problem, leading to larger variance. This issue can be mitigated through ``sample accumulation'', accumulating samples across multiple consecutive steps, similar to gradient accumulation practice in LLMs. A pseudo-code for sample accumulation can be found in Appendix~\ref{sec:milp}. In addition, though integer linear programming is NP-hard~\citep{vazirani1997approximation}, the number of variables in our batch-level optimization is typically small, resulting in minimal computational overhead. Refer to the wall-time plot for the SciPy solver in Appendix~\ref{sec:milp} for further details.

{\bf Controlled setting.} We keep our experiment setting minimal: we do not use reward models; we use only an 8B model to generate any set of sampled responses, $\mathbb{A}^\textsc{sample}$. Our RL training set, $\mathbb{D}_\textsc{rl}$, contains only 10k prompts, leaving the setting quite controlled and providing room for improvement through engineering efforts.

{\bf Limitations.} Our work is limited in its choice of RL algorithms and applications. While it should be straightforward to apply our framework to different RL frameworks, additional effort is required to derive an optimization problem for different application tasks. Future work may include extending our constrained policy optimization framework to a broader range of LLM applications or scaling up our experiments.

{\bf Broader applications.} 
Although our framework has only been evaluated with inference budget allocation, the resource allocation problem has far-reaching implications within the ML community. As a result, our framework can be potentially extended to further LLM applications. 
For example, a promising application is statistical parity~\citep{zemel2013learning}, aka group fairness. 
In this context, one could consider attributing responses to their respective social groups, and cap the density of responses that correspond to socially privileged groups, to encourage more inclusive and equitable responses across different demographics.
Another potential application is the balanced activation of mixture of experts~\citep{jacobs1991adaptive, shazeer2017outrageously, lepikhin2020gshard}, which is sometimes achieved by adding an auxiliary loss~\citep{wei2024skywork}. Alternatively, this balance can be possibly achieved by enforcing a minimal total activation density for each expert. This may help to prevent over-reliance on a subset of experts, and hence enhancing the overall robustness and efficiency.
We leave the exploration of broader applications and their implementations as future directions.

%% file: appendix/apx_responses.tex
\section{Sample Responses}

We show a couple of example responses of ASV-IuB-$50\%$, the omitted prompt could be found in Figure~\ref{fig:prompts}.

\begin{tcolorbox}[title=\textbf{Voting Example}]
\footnotesize

{\color{RoyalBlue} Prompt omitted ...}
~\\

Problem: Compute $\arcsin \left( -\frac{1}{2} \right).$  Express your answer in radians.

~\\

[TRIAL] \#\# Step 1: Recall the definition of arcsine
The function $\arcsin x$ is the inverse of the sine function. It returns the angle whose sine is $x$.

\#\# Step 2: Use the unit circle to find the angle
From the unit circle, we know that $\sin \left( -\frac{\pi}{3} \right) = -\frac{1}{2}$.

\#\# Step 3: Determine the value of $\arcsin \left( -\frac{1}{2} \right)$
Since $\sin \left( -\frac{\pi}{3} \right) = -\frac{1}{2}$, we have $\arcsin \left( -\frac{1}{2} \right) = -\frac{\pi}{3}$.

The final answer is: $\boxed{-\frac{\pi}{3}}$.
[/TRIAL] 

[TRIAL] \#\# Step 1: Recall the sine function values for common angles.
The sine of $\frac{\pi}{3}$ is $\frac{\sqrt{3}}{2}$.

\#\# Step 2: Use the relationship between sine and arcsine.
Given $\sin(\theta) = -\frac{1}{2}$, we are looking for $\arcsin\left(-\frac{1}{2}\right)$.

\#\# Step 3: Identify the angle whose sine is $-\frac{1}{2}$.
From the unit circle or trigonometric identities, we know that $\sin\left(-\frac{\pi}{6}\right) = -\frac{1}{2}$.

\#\# Step 4: Express the answer in radians.
Therefore, $\arcsin\left(-\frac{1}{2}\right) = -\frac{\pi}{6}$.

Therefore, the final answer is: $\boxed{-\frac{\pi}{6}}$
[/TRIAL] 

[TRIAL] \#\# Step 1: Recall the sine function's relationship with the unit circle.
The sine of an angle in a right triangle is the ratio of the length of the side opposite the angle to the length of the hypotenuse. For angles in the unit circle (a circle with radius 1), the sine of an angle is the y-coordinate of the point where the terminal side of the angle intersects the circle.

\#\# Step 2: Identify the angle with a sine of $-\frac{1}{2}$.
From the unit circle, we know that $\sin\left(\frac{\pi}{3}\right) = \frac{1}{2}$. Since sine is negative in the third and fourth quadrants, and we are looking for an angle whose sine is $-\frac{1}{2}$, the angle must be in either the third or fourth quadrant.

\#\# Step 3: Determine the angle in the correct quadrant.
Considering the unit circle, the reference angle for $\frac{\pi}{3}$ is $\frac{\pi}{3}$ itself. Since we are looking for an angle with a sine of $-\frac{1}{2}$, and knowing that sine is negative in the third and fourth quadrants, the angle we are looking for is in the fourth quadrant because $\sin\left(\frac{5\pi}{3}\right)$ would indeed be negative.

\#\# Step 4: Calculate the angle.
The angle whose sine is $-\frac{1}{2}$ is $\frac{5\pi}{3}$.

Therefore, the final answer is: $\boxed{\frac{5\pi}{3}}$
[/TRIAL] 

[TRIAL] \#\# Step 1: Recall the definition of arcsine
The function $\arcsin x$ is the inverse of the sine function. It returns the angle whose sine is $x$.

\#\# Step 2: Use the sine function to find the angle
We want to find the angle whose sine is $-\frac{1}{2}$. We can use the sine function to find this angle.

\#\# Step 3: Recall common angles and their sines
From the unit circle or trigonometric identities, we know that $\sin\left(\frac{\pi}{6}\right) = \frac{1}{2}$. Since sine is an odd function, $\sin\left(-\frac{\pi}{6}\right) = -\frac{1}{2}$.

\#\# Step 4: Determine the angle
Therefore, the angle whose sine is $-\frac{1}{2}$ is $-\frac{\pi}{6}$.

Therefore, the final answer is: $\boxed{-\frac{\pi}{6}}$
[/TRIAL] 

[TRIAL] \#\# Step 1: Recall the sine function's values for common angles.
The sine function has well-known values for certain angles. Specifically, $\sin\left(\frac{\pi}{6}\right) = \frac{1}{2}$.

\#\# Step 2: Use the symmetry and periodicity of the sine function.
Given that $\sin\left(\frac{\pi}{6}\right) = \frac{1}{2}$, and knowing that sine is an odd function, we have $\sin\left(-\frac{\pi}{6}\right) = -\frac{1}{2}$. This step utilizes the property of odd functions where $f(-x) = -f(x)$.

\#\# Step 3: Solve for $\arcsin \left( -\frac{1}{2} \right)$.
Given that $\sin\left(-\frac{\pi}{6}\right) = -\frac{1}{2}$, the angle whose sine is $-\frac{1}{2}$ is $-\frac{\pi}{6}$. This is because the inverse sine function, $\arcsin$, returns the angle whose sine is a given value.

Therefore, the final answer is: $\boxed{-\frac{\pi}{6}}$
[/TRIAL] 

The answer $\boxed{-\frac{\pi}{6}}$ has occured three times and is considered as a consensus.
The final answer is $\boxed{-\frac{\pi}{6}}$. I hope it is correct.

\end{tcolorbox}

\begin{tcolorbox}[title=\textbf{Non-voting Example}]
\footnotesize

{\color{RoyalBlue} Prompt omitted ...}
~\\

Problem: If $f(x) = \frac{3x-2}{x-2}$, what is the value of $f(-2) +f(-1)+f(0)$? Express your answer as a common fraction.

~\\

[TRIAL] \#\# Step 1: Substitute $x=-2$ into the function $f(x)$.
To find $f(-2)$, substitute $x=-2$ into the function:
\[f(-2) = \frac{3(-2)-2}{-2-2} = \frac{-6-2}{-4} = \frac{-8}{-4} = 2.\]

\#\# Step 2: Substitute $x=-1$ into the function $f(x)$.
To find $f(-1)$, substitute $x=-1$ into the function:
\[f(-1) = \frac{3(-1)-2}{-1-2} = \frac{-3-2}{-3} = \frac{-5}{-3} = \frac{5}{3}.\]

\#\# Step 3: Substitute $x=0$ into the function $f(x)$.
To find $f(0)$, substitute $x=0$ into the function:
\[f(0) = \frac{3(0)-2}{0-2} = \frac{0-2}{-2} = \frac{-2}{-2} = 1.\]

\#\# Step 4: Sum the values of $f(-2)$, $f(-1)$, and $f(0)$.
Sum the calculated values:
\[f(-2) + f(-1) + f(0) = 2 + \frac{5}{3} + 1.\]

\#\# Step 5: Simplify the sum to express it as a common fraction.
To combine the terms into a common fraction, convert the whole numbers to fractions with a common denominator:
\[2 + \frac{5}{3} + 1 = \frac{6}{3} + \frac{5}{3} + \frac{3}{3} = \frac{6 + 5 + 3}{3} = \frac{14}{3}.\]

Therefore, the value of $f(-2) + f(-1) + f(0)$ is $\frac{14}{3}$.

Therefore, the final answer is: $\boxed{\frac{14}{3}}$
[/TRIAL]

Terminated due to difficulty level.

\end{tcolorbox}

%% file: appendix/apx_accumulation.tex
\section{Batch Accumulation}

This is a batch accumulation implementation of Algorithm~\ref{algo:cgpo}. For brevity, we use $\textsc{opt}_\textsc{iub}$ as an example. One could increase the optimization problem size of $n \times m$ to $(n\cdot k_b) \times (m \cdot k_r)$ using Algorithm~\ref{algo:cgpo-accumulation}, where superscripts indicate matrix shape, left subscripts denote accumulation indices (distinguishing them from element indices).

\begin{algorithm} 
\caption{IBPO with Sample Accumulation}
\begin{algorithmic}[1]
\REQUIRE 
prompt set $\mathbb{D}$, 
batch size $n$, 
number of responses $m$,  
init policy $\pi_0 = \pi_\mathrm{ref}$, 
num of iters $T$, 
budgets $q_+$
\FOR{$t = 1, \ldots, T$}
    \FOR{$i = 1, \ldots, k_b$} 
        \STATE {\bf prompt sampling}: $ \prescript{}{i}{\xmat}^n \sim \mu_\mathbb{D}$ 
        \FOR{$j = 1, \ldots, k_r$} 
            \STATE {\bf response generation}: $ \prescript{}{ij}{\ymat}^{n\times m} \sim \pi_{\theta_t}(\prescript{}{i}{\xmat})$
        \ENDFOR
    \ENDFOR
    \STATE {\bf prompt accumulation}: $\tilde{\xmat}^{(n \cdot k_b)} = [\prescript{}{1}{\xmat}, \prescript{}{2}{\xmat}, \cdots, \prescript{}{k_b}{\xmat}]$
    \STATE {\bf response accumulation}: 
    $\tilde{\ymat}^{(n \cdot k_b) \times (m \cdot k_r)} = [\prescript{}{11}{\ymat}, \prescript{}{12}{\ymat}, \cdots, \prescript{}{1 k_r}{\ymat}; \cdots; \prescript{}{k_b 1}{\ymat}, \prescript{}{k_b 2}{\ymat}, \cdots, \prescript{}{k_b k_r}{\ymat}]$
    \STATE {\bf evaluate correctness} and {\bf empirical KL}: $\rmat_\mathrm{match} = r_\mathrm{match}(\tilde{\xmat}, \tilde{\ymat})$ and $\hat{\mathbf{KL}} = \hat{\KL}(\tilde{\ymat}; \tilde{\xmat}, \pi_\mathrm{ref})$
    \STATE {\bf margin maximization}: $\hat{\pi}^\star_{\tilde{\xmat}, \tilde{\ymat}} \in  \textsc{opt}_\textsc{iub}(\tilde{\xmat}, \tilde{\ymat}, \rmat_\mathrm{match}, \hat{\mathbf{KL}}, q_+)$ as defined in Eq.~\eqref{eq:final}
    \STATE {\bf gradient update}: with
    $-\sum_{i=1}^{n\cdot k_b}\sum_{j=1}^{m\cdot k_r}
    {\hat{\pi}^\star_{\tilde{\xmat}, \tilde{\ymat}}(y_{ij} | x_i)}
    \frac{\partial}{\partial \theta}
    \log \pi_\theta(y_{ij} | x_i) \big|_{\theta = \theta_t} $
\ENDFOR
\end{algorithmic}\label{algo:cgpo-accumulation}
\end{algorithm}

\newpage

%% file: appendix/apx_solver.tex
\section{MILP Solving}\label{sec:milp}

\begin{figure}[h] 
    \centering
    \centering
    \includegraphics[width=0.4\textwidth]{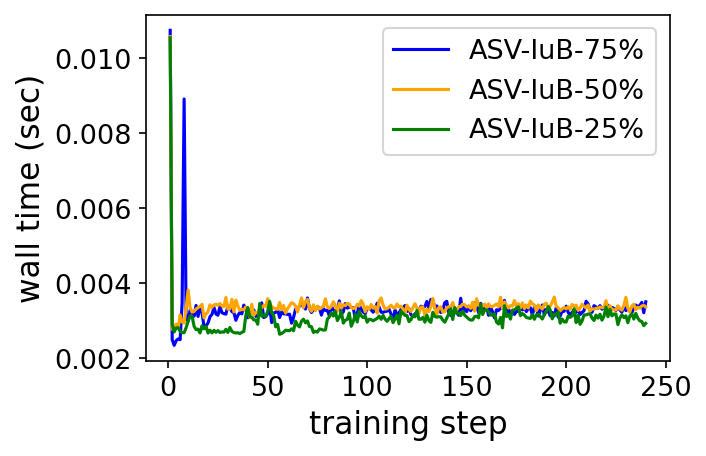}
    \caption{Wall time (averaged across ranks) spent by solver in seconds.}
    \label{fig:wall-time}
\end{figure}

We use SciPy MILP solver, available \href{https://docs.scipy.org/doc/scipy/reference/generated/scipy.optimize.milp.html}{here}, to solve our in-batch integer linear programming optimizations. A Pythonic pseudo code is available in Figure~\ref{lst:milp}. And the wall-time consumed could be found in Figure~\ref{fig:wall-time}. In general, the problem size is small, so the computational overhead is negligible.

\begin{lstlisting}[caption={Pythonic code snippet for solving IuB with SciPy: Note that this is for demonstration purposes, and error-free execution is not guaranteed, due to omitted corner cases.}, label={lst:milp}, language=Python]
import numpy as np
from scipy.optimize import milp, LinearConstraint, Bounds

def solve_iub(acceptance, is_vote, budget):
    """ solves a Inference under Budget (IuB) problem.
    parameters:
    - acceptance: an n x m array where each element is 1 if accepted, 0 otherwise.
    - is_vote: an n x m array indicating whether response is voting (1) or non-voting (0).
    - budget: the fractional budget (q+) constraint for the problem. """
    n, m = acceptance.shape
    # calculate pass rates for vote-based and non-vote-based responses
    vote_pass_rate = np.mean(acceptance * is_vote, axis=1, keepdims=True)
    sample_pass_rate = np.mean(acceptance * (1 - is_vote), axis=1, keepdims=True)
    margin = vote_pass_rate - sample_pass_rate  # (n, 1)
    # flattern acceptance and vote indicator, tile the margin
    acceptance = np.reshape(acceptance, -1)             # (n x m, )
    is_vote = np.reshape(is_vote, -1)                   # (n x m, )
    margin = np.reshape(np.tile(margin, [1, m]), -1)    # (n x m, )
    
    # define the objective function coefficients
    c = -1 * margin * is_vote + margin * (1 - is_vote)
    # acceptance constraints: ensure each prompt meets acceptance criteria
    A_acceptance = np.eye(len(acceptance))
    b_acceptance = acceptance
    # one response per problem constraint (BoN)
    A_problem = np.zeros((n, n * m))
    for i in range(n):
        A_problem[i, i * m:(i + 1) * m] = 1
    b_problem = np.ones(n)
    # voting responses budget constraint
    A_vote_budget = np.where(is_vote == 1, 1, 0).reshape(1, -1)
    vote_budget = np.round(budget * len(acceptance))
    
    # combine all constraints into a single matrix
    A = np.vstack([A_acceptance, A_problem, A_vote_budget])
    b_lower = -np.inf * np.ones(A.shape[0])  # lower bounds for constraints
    b_upper = np.hstack([b_acceptance, b_problem, vote_budget])  # upper bounds 
    # solve the MILP problem using the defined objective and constraints
    result = milp(c, integrality=np.ones(len(c)), bounds=Bounds(0, 1),
                  constraints=LinearConstraint(A, b_lower, b_upper))
    return result
\end{lstlisting}

%% file: appendix/apx_data.tex
\section{Data Construction}\label{sec:data}

We elaborate on the details of $\mathbb{Q}_\textsc{sdpo}$, $\mathbb{A}_\textsc{sdpo}^\textsc{golden}$, $\mathbb{Q}_\textsc{math}$, $\mathbb{A}_\textsc{math}^\textsc{sample}$, and the filteration function $\mathcal{F}$. 

{\bf SDPO dataset.} \citet{lai2024step} create a dataset (see their Section 3.2) contains 10,795 step-wise preference pairs, where the winning response is a correct solution, the losing response is incorrect starting from certain step. For our purpose of SFT and RL training, we do not need losing responses, hence we create $\mathbb{A}_\textsc{sdpo}^\textsc{golden}$ with only wining (correct) responses. The step-formatted response (see an example below) is the reason we choose the dataset from \citet{lai2024step}, as it adhere the SCoT format of LLaMA models. Therefore, one do not need to worry about potential format changes.

\begin{tcolorbox}[title=\textbf{SDPO question-response example}]
\footnotesize

Problem: Sides of a rectangular park are in the ratio 3: 2 and its area is 3750 sq m, the cost of fencing it at 40 ps per meter is?

~\\

\#\# Step 1:  Let's assume the sides of the rectangle are $3x$ and $2x$.
The area of the rectangle is the product of its sides, so we have $3x \cdot 2x = 6x^2 = 3750$.
Solving for $x$, we get $x^2 = \frac{3750}{6} = 625$, so $x = \sqrt{625} = 25$.

\#\# Step 2: So the sides of the rectangle are $3x = 3 \cdot 25 = 75$ meters and $2x = 2 \cdot 25 = 50$ meters.

\#\# Step 3: The perimeter of the rectangle is $2 \times (75 + 50) = 2 \times 125 = 250$ meters.

\#\# Step 4: The cost of fencing is $40$ paise per meter, so the total cost of fencing is $250 \times 40 = 10000$ paise.
Since $1$ rupee is equal to $100$ paise, the cost in rupees is $\frac{10000}{100} = 100$ rupees.

Therefore, the final answer is: $\boxed{100}$

\end{tcolorbox}

{\bf SV dataset for MATH.} We also created datasets, $\mathbb{D}_\textsc{sv}$ and $\mathbb{D}_\textsc{asv1}$, for SFT so that a model could follow the SV instructions. We take $\mathbb{D}_\textsc{sv}$ as an example and $\mathbb{D}_\textsc{asv1}$ could be created similarly. To do so, we first generate $32$ responses per prompt for the entire MATH training split with a temperature of $1.2$ and top-p of $0.9$. We then apply our SV templates $\mathcal{T}_q$ and $\mathcal{T}_a$ to create corresponding SV question and answer pairs. The procedure of creating a SV response is given by Algorithm~\ref{algo:creating-sv}. While we have include an example of SV responses in Figure~\ref{fig:sv-templates}, we make it more concrete the SV response template below,

\begin{figure}[!h]
\begin{minipage}{0.5\textwidth}

\begin{tcolorbox}[title=\textbf{SV template: $\mathcal{T}_a(A_1, \cdots, A_k, \text{final answer})$ (if consensus found)}]

\small 

[TRIAL] \{$A_1$\} [/TRIAL]

[TRIAL] \{$A_2$\} [/TRIAL]

$\cdots$ 

[TRIAL] \{$A_k$\} [/TRIAL] 

The answer $\boxed{\text{final answer}}$ has occurred three times, and is considered as a consensus.

The final answer is $\boxed{\text{final answer}}$. I hope it is correct.

\end{tcolorbox}

\end{minipage}
\hfill
\begin{minipage}{0.5\textwidth}

\begin{tcolorbox}[title=\textbf{SV template: $\mathcal{T}_a(A_1, \cdots, A_8, \text{final answer})$ (if no consensus found)}]

\small 

[TRIAL] \{$A_1$\} [/TRIAL]

[TRIAL] \{$A_2$\} [/TRIAL]

$\cdots$

[TRIAL] \{$A_8$\} [/TRIAL] 

Maximum trials reached but no consensus found due to a tie; the most plausible answer is $\boxed{\text{final answer}}$.

The final answer is $\boxed{\text{final answer}}$. I hope it is correct.

\end{tcolorbox}

\end{minipage}
\caption{SV response templates. Left: suppose a consensus is found at $k$-th answer; Right: no consensus found. Note the subscript $i$ of $A_i$ only denotes index of answer, $A_i$ and $A_j$ could still have same final answer for $i\neq j$.}
\label{fig:sv-templates}
\end{figure}

\begin{algorithm} 
\caption{Creating SV Response for SFT .}
\begin{algorithmic}[1]
\REQUIRE template $\mathcal{T}_q$, $\mathcal{T}_a$, a problem $Q$, a set of shuffled responses $\{A_i: i=1, 2, \cdots, K\}$
\STATE create prompt: $\mathcal{T}_q(Q)$ \hfill \textcolor{gray}{replace problem placeholder with $Q$} 
\STATE responses = []; final\_answer = [INVALID\_ANSWER]; found = False
\FOR{$i = 1, 2, \cdots, K$}
    \STATE responses.append($A_i$)
    \STATE majority = find\_majority(responses)
    \STATE majority\_count = count\_majority(responses, majority)
    \IF{majority\_count == 3}
        \STATE found = True; final\_answer = majority; break
    \ENDIF
\ENDFOR
\IF{found == False {\bf and} responses contain correct solution}
    \STATE final\_answer = random pick a correct solution 
\ENDIF
\STATE create SV response: $\mathcal{T}_a(\text{responses, final\_answer})$
\end{algorithmic}\label{algo:creating-sv}
\end{algorithm}

\begin{figure}[!t]
    \centering
    \begin{subfigure}[b]{0.485\textwidth}
        \includegraphics[width=\textwidth]{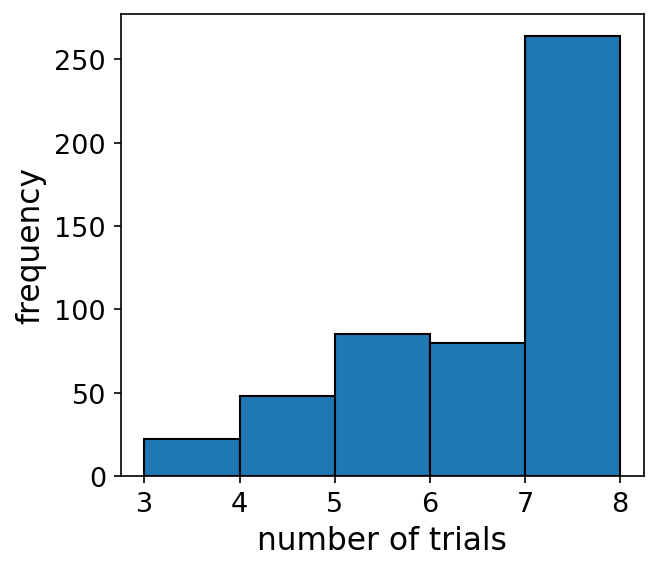}
        \caption{Distribution of number of trials per response}
        \label{fig:trial-distr}
    \end{subfigure}
    \hfill
    \begin{subfigure}[b]{0.485\textwidth}
        \includegraphics[width=\textwidth]{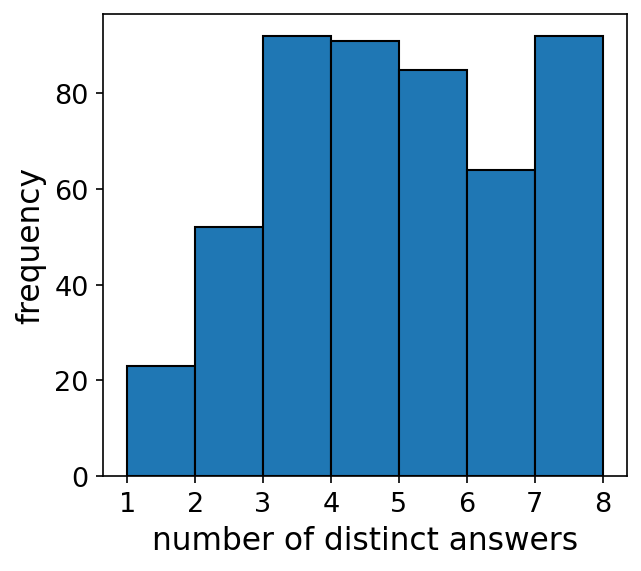}
        \caption{Distribution of number of distinct answers}
        \label{fig:answer-distr}
    \end{subfigure}
    \caption{The distribution of filtered prompt-responses subset, which suggests that the construct data is generally diversely distributed.}
    \label{fig:sv-distr}
\end{figure}

We then apply some filtration $\mathcal{F}$ to the created dataset. 
We remove any SV responses where the final answers are incorrect. From the remaining set, we sub-sample 500 question-response pairs. These pairs are selected from problems that have between 4 and 8 distinct answers out of 32 samples, ensuring that we construct sequential responses with a diverse trials.
The distribution of trial counts and distinct answer counts are shown in Figure~\ref{fig:sv-distr}. It can be observed that the filtered data is generally diversely distributed.

%% file: appendix/hyperparams.tex
\newpage
~\newpage
\section{Hyperparameters}\label{sec:hyperparams}

We list the hyperparameters used for experiments setup 1.2, 2.1, 2.2, as described in Table~\ref{tab:pipelines}. And we conduct our experiments with NVIDIA-A100-80Gs. (Please refer to~\citet{xu2024perfect} for the definition of some RL-specific hyper-parameters.)

\begin{table}[h] 
    \centering
    \caption{Hyperparameters for Experiment setups 1.2, 2.1, and 2.2}
    \label{tab:hyperparameters}
    \small
    \begin{tabular}{l|lll}
        \toprule
        \textbf{Hyperparameter} & \textbf{Setup 1.2} & \textbf{Setup 2.1} & \textbf{Setup 2.2} \\
        \midrule
        prompt size            & 11295 & 11295 & 10795  \\
        number of nodes        & 4                 & 4                & 8               \\
        learning rate          & 1e-6              & 1e-6             & 5e-7               \\
        batch size (per node)  & 8                 & 16               & 4                 \\
        num of steps           & 1024              & 2048             & 240                \\
        optimizer              & AdamW & AdamW & AdamW \\
        scheduler              & constant          & constant         & constant               \\
        packing                & yes               & yes  & - \\
        max sequence length    & 32768             & 32768             & 6144                \\
        gradient accumulation  & 1                 & 2                 & 1                  \\
        \midrule
        \multicolumn{4}{c}{RL-specific params} \\
        \midrule
        num generation per prompt &                &                  & 8 \\
        max generation length  &                   &                  & 4096 \\
        temperature & & & 1.0 \\
        top-p & & & 0.9 \\ 
        KL-threashold & & & 1024 \\
        batch accumulation $k_b$ & & & 4 \\
        response accumulation $k_r$ & & & 1 \\
        \bottomrule
    \end{tabular}
\end{table}

\normalsize

%% file: appendix/apx_budget_following.tex
\section{Further Discussions}\label{sec:further-discussion}

\begin{figure}[h] 
    \centering
    \begin{subfigure}[b]{0.485\textwidth}
        \includegraphics[width=\textwidth]{figures/budget-train-passed.png}
        \caption{voting ratio of correct responses (training set)}
        \label{fig:budgets-train-passed}
    \end{subfigure}
    \begin{subfigure}[b]{0.485\textwidth}
        \includegraphics[width=\textwidth]{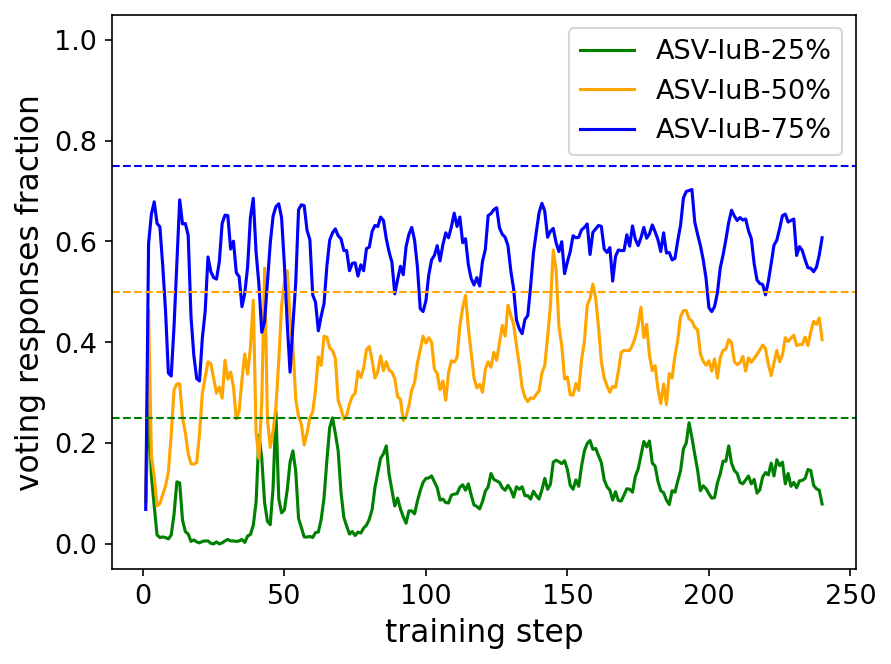}
        \caption{voting ratio of all responses (training set)}
        \label{fig:budgets-train-all}
    \end{subfigure}
    \caption{Voting response ratio versus training steps. Dashed line denotes the budget $q_+$. On the training set, IuB formulation follows the budget constraints almost exactly for both: (a) correct responses; (b) all responses.}
    \label{fig:budgets-apx}
\end{figure}

{\bf Budgets constraints.} While in Section~\ref{sec:exp}, we show that our formulation follows the constraints exactly on the training set, for responses that are correct. Figure~\ref{fig:budgets-train-all} further shows the ratio of voting responses for all online rollouts. It is clear that the constraints are met as well.

\newpage

\begin{figure}[h]
    \centering
    \includegraphics[width=0.485\textwidth]{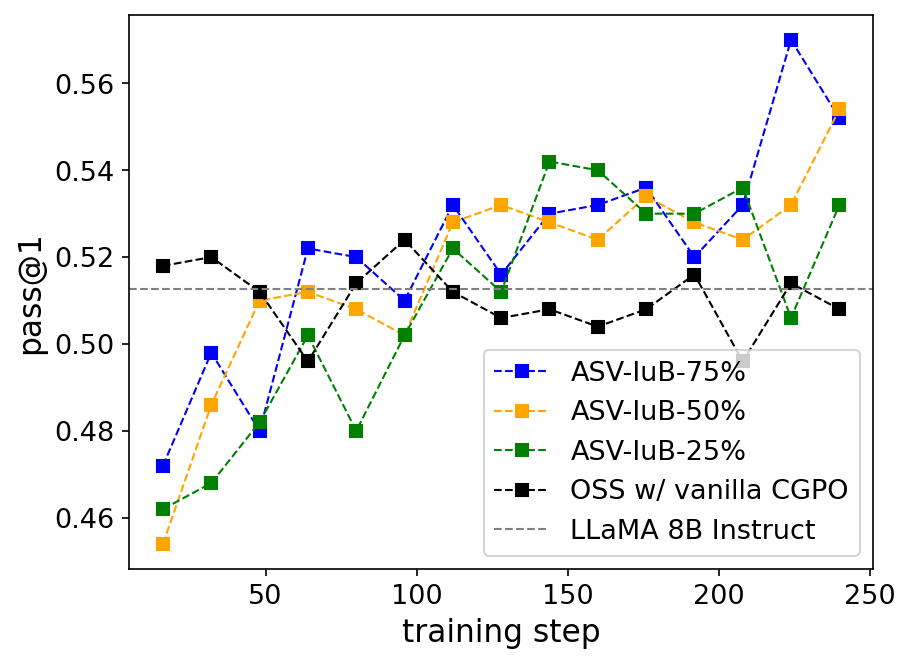}
    \caption{Training curves: each point corresponds an evaluation of MATH500 test set, and dashed line is the pass@$1$ of LLaMA 8B Instruct.}
    \label{fig:training-curves}
\end{figure}

{\bf CGPO on $\mathbb{D}_\textsc{rl}$ w/ LLaMA.} To further understand the results we presented in Section~\ref{sec:exp} is benfit from our budget-aware formulation or from the prompt set of~\citet{lai2024step}. We further run CGPO with open-sourced LLaMA model and the SDPO dataset in the SCoT format. Figure~\ref{fig:training-curves} compares the training dynamics of CGPO with LLaMA versus our ASU-IuB-$q_+$ experiments. In general, the SDPO prompt set does not provide much additional knowledge as suggested by the OSS w/ vanilla CGPO experiment, but ASV-IuB-$q_+$ experiments are able to achieve noticable gain.

%% file: paper.bbl
\begin{thebibliography}{82}
\providecommand{\natexlab}[1]{#1}
\providecommand{\url}[1]{\texttt{#1}}
\expandafter\ifx\csname urlstyle\endcsname\relax
  \providecommand{\doi}[1]{doi: #1}\else
  \providecommand{\doi}{doi: \begingroup \urlstyle{rm}\Url}\fi

\bibitem[Aggarwal et~al.(2023)Aggarwal, Madaan, Yang, et~al.]{aggarwal2023let}
Pranjal Aggarwal, Aman Madaan, Yiming Yang, et~al.
\newblock Let's sample step by step: Adaptive-consistency for efficient reasoning and coding with llms.
\newblock \emph{arXiv preprint arXiv:2305.11860}, 2023.

\bibitem[Altman(2021)]{altman2021constrained}
Eitan Altman.
\newblock \emph{Constrained Markov decision processes}.
\newblock Routledge, 2021.

\bibitem[Amos and Kolter(2017)]{amos2017optnet}
Brandon Amos and J~Zico Kolter.
\newblock Optnet: Differentiable optimization as a layer in neural networks.
\newblock In \emph{International conference on machine learning}, pages 136--145. PMLR, 2017.

\bibitem[Anthony et~al.(2020)Anthony, Kanding, and Selvan]{anthony2020carbontracker}
Lasse F~Wolff Anthony, Benjamin Kanding, and Raghavendra Selvan.
\newblock Carbontracker: Tracking and predicting the carbon footprint of training deep learning models.
\newblock \emph{arXiv preprint arXiv:2007.03051}, 2020.

\bibitem[Austin et~al.(2021)Austin, Odena, Nye, Bosma, Michalewski, Dohan, Jiang, Cai, Terry, Le, et~al.]{austin2021program}
Jacob Austin, Augustus Odena, Maxwell Nye, Maarten Bosma, Henryk Michalewski, David Dohan, Ellen Jiang, Carrie Cai, Michael Terry, Quoc Le, et~al.
\newblock Program synthesis with large language models.
\newblock \emph{arXiv preprint arXiv:2108.07732}, 2021.

\bibitem[Badanidiyuru et~al.(2018)Badanidiyuru, Kleinberg, and Slivkins]{badanidiyuru2018bandits}
Ashwinkumar Badanidiyuru, Robert Kleinberg, and Aleksandrs Slivkins.
\newblock Bandits with knapsacks.
\newblock \emph{Journal of the ACM (JACM)}, 65\penalty0 (3):\penalty0 1--55, 2018.

\bibitem[Bubeck et~al.(2015)Bubeck, Dekel, Koren, and Peres]{bubeck2015bandit}
S{\'e}bastien Bubeck, Ofer Dekel, Tomer Koren, and Yuval Peres.
\newblock Bandit convex optimization:$\backslash$sqrtt regret in one dimension.
\newblock In \emph{Conference on Learning Theory}, pages 266--278. PMLR, 2015.

\bibitem[Cayci et~al.(2022)Cayci, Zheng, and Eryilmaz]{cayci2022lyapunov}
Semih Cayci, Yilin Zheng, and Atilla Eryilmaz.
\newblock A lyapunov-based methodology for constrained optimization with bandit feedback.
\newblock In \emph{Proceedings of the AAAI Conference on Artificial Intelligence}, volume~36, pages 3716--3723, 2022.

\bibitem[Chen et~al.(2021)Chen, Tworek, Jun, Yuan, Pinto, Kaplan, Edwards, Burda, Joseph, Brockman, et~al.]{chen2021evaluating}
Mark Chen, Jerry Tworek, Heewoo Jun, Qiming Yuan, Henrique Ponde De~Oliveira Pinto, Jared Kaplan, Harri Edwards, Yuri Burda, Nicholas Joseph, Greg Brockman, et~al.
\newblock Evaluating large language models trained on code.
\newblock \emph{arXiv preprint arXiv:2107.03374}, 2021.

\bibitem[Chen and He(2021)]{chen2021exploring}
Xinlei Chen and Kaiming He.
\newblock Exploring simple siamese representation learning.
\newblock In \emph{Proceedings of the IEEE/CVF conference on computer vision and pattern recognition}, pages 15750--15758, 2021.

\bibitem[Chenery and Kretschmer(1956)]{chenery1956resource}
Hollis~B Chenery and Kenneth~S Kretschmer.
\newblock Resource allocation for economic development.
\newblock \emph{Econometrica, Journal of the Econometric Society}, pages 365--399, 1956.

\bibitem[Chentanez et~al.(2004)Chentanez, Barto, and Singh]{chentanez2004intrinsically}
Nuttapong Chentanez, Andrew Barto, and Satinder Singh.
\newblock Intrinsically motivated reinforcement learning.
\newblock \emph{Advances in neural information processing systems}, 17, 2004.

\bibitem[Chow et~al.(2018)Chow, Nachum, Duenez-Guzman, and Ghavamzadeh]{chow2018lyapunov}
Yinlam Chow, Ofir Nachum, Edgar Duenez-Guzman, and Mohammad Ghavamzadeh.
\newblock A lyapunov-based approach to safe reinforcement learning.
\newblock \emph{Advances in neural information processing systems}, 31, 2018.

\bibitem[Cobbe et~al.(2021)Cobbe, Kosaraju, Bavarian, Chen, Jun, Kaiser, Plappert, Tworek, Hilton, Nakano, et~al.]{cobbe2021training}
Karl Cobbe, Vineet Kosaraju, Mohammad Bavarian, Mark Chen, Heewoo Jun, Lukasz Kaiser, Matthias Plappert, Jerry Tworek, Jacob Hilton, Reiichiro Nakano, et~al.
\newblock Training verifiers to solve math word problems.
\newblock \emph{arXiv preprint arXiv:2110.14168}, 2021.

\bibitem[Cplex(2009)]{cplex2009v12}
IBM~ILOG Cplex.
\newblock V12. 1: User’s manual for cplex.
\newblock \emph{International Business Machines Corporation}, 46\penalty0 (53):\penalty0 157, 2009.

\bibitem[DeepSeek-AI et~al.(2025)DeepSeek-AI, Guo, Yang, Zhang, Song, Zhang, Xu, Zhu, Ma, Wang, Bi, Zhang, Yu, Wu, Wu, Gou, Shao, Li, Gao, Liu, Xue, Wang, Wu, Feng, Lu, Zhao, Deng, Zhang, Ruan, Dai, Chen, Ji, Li, Lin, Dai, Luo, Hao, Chen, Li, Zhang, Bao, Xu, Wang, Ding, Xin, Gao, Qu, Li, Guo, Li, Wang, Chen, Yuan, Qiu, Li, Cai, Ni, Liang, Chen, Dong, Hu, Gao, Guan, Huang, Yu, Wang, Zhang, Zhao, Wang, Zhang, Xu, Xia, Zhang, Zhang, Tang, Li, Wang, Li, Tian, Huang, Zhang, Wang, Chen, Du, Ge, Zhang, Pan, Wang, Chen, Jin, Chen, Lu, Zhou, Chen, Ye, Wang, Yu, Zhou, Pan, Li, Zhou, Wu, Ye, Yun, Pei, Sun, Wang, Zeng, Zhao, Liu, Liang, Gao, Yu, Zhang, Xiao, An, Liu, Wang, Chen, Nie, Cheng, Liu, Xie, Liu, Yang, Li, Su, Lin, Li, Jin, Shen, Chen, Sun, Wang, Song, Zhou, Wang, Shan, Li, Wang, Wei, Zhang, Xu, Li, Zhao, Sun, Wang, Yu, Zhang, Shi, Xiong, He, Piao, Wang, Tan, Ma, Liu, Guo, Ou, Wang, Gong, Zou, He, Xiong, Luo, You, Liu, Zhou, Zhu, Xu, Huang, Li, Zheng, Zhu, Ma, Tang, Zha, Yan, Ren, Ren, Sha, Fu, Xu, Xie, Zhang,
  Hao, Ma, Yan, Wu, Gu, Zhu, Liu, Li, Xie, Song, Pan, Huang, Xu, Zhang, and Zhang]{deepseekai2025deepseekr1incentivizingreasoningcapability}
DeepSeek-AI, Daya Guo, Dejian Yang, Haowei Zhang, Junxiao Song, Ruoyu Zhang, Runxin Xu, Qihao Zhu, Shirong Ma, Peiyi Wang, Xiao Bi, Xiaokang Zhang, Xingkai Yu, Yu~Wu, Z.~F. Wu, Zhibin Gou, Zhihong Shao, Zhuoshu Li, Ziyi Gao, Aixin Liu, Bing Xue, Bingxuan Wang, Bochao Wu, Bei Feng, Chengda Lu, Chenggang Zhao, Chengqi Deng, Chenyu Zhang, Chong Ruan, Damai Dai, Deli Chen, Dongjie Ji, Erhang Li, Fangyun Lin, Fucong Dai, Fuli Luo, Guangbo Hao, Guanting Chen, Guowei Li, H.~Zhang, Han Bao, Hanwei Xu, Haocheng Wang, Honghui Ding, Huajian Xin, Huazuo Gao, Hui Qu, Hui Li, Jianzhong Guo, Jiashi Li, Jiawei Wang, Jingchang Chen, Jingyang Yuan, Junjie Qiu, Junlong Li, J.~L. Cai, Jiaqi Ni, Jian Liang, Jin Chen, Kai Dong, Kai Hu, Kaige Gao, Kang Guan, Kexin Huang, Kuai Yu, Lean Wang, Lecong Zhang, Liang Zhao, Litong Wang, Liyue Zhang, Lei Xu, Leyi Xia, Mingchuan Zhang, Minghua Zhang, Minghui Tang, Meng Li, Miaojun Wang, Mingming Li, Ning Tian, Panpan Huang, Peng Zhang, Qiancheng Wang, Qinyu Chen, Qiushi Du, Ruiqi Ge, Ruisong
  Zhang, Ruizhe Pan, Runji Wang, R.~J. Chen, R.~L. Jin, Ruyi Chen, Shanghao Lu, Shangyan Zhou, Shanhuang Chen, Shengfeng Ye, Shiyu Wang, Shuiping Yu, Shunfeng Zhou, Shuting Pan, S.~S. Li, Shuang Zhou, Shaoqing Wu, Shengfeng Ye, Tao Yun, Tian Pei, Tianyu Sun, T.~Wang, Wangding Zeng, Wanjia Zhao, Wen Liu, Wenfeng Liang, Wenjun Gao, Wenqin Yu, Wentao Zhang, W.~L. Xiao, Wei An, Xiaodong Liu, Xiaohan Wang, Xiaokang Chen, Xiaotao Nie, Xin Cheng, Xin Liu, Xin Xie, Xingchao Liu, Xinyu Yang, Xinyuan Li, Xuecheng Su, Xuheng Lin, X.~Q. Li, Xiangyue Jin, Xiaojin Shen, Xiaosha Chen, Xiaowen Sun, Xiaoxiang Wang, Xinnan Song, Xinyi Zhou, Xianzu Wang, Xinxia Shan, Y.~K. Li, Y.~Q. Wang, Y.~X. Wei, Yang Zhang, Yanhong Xu, Yao Li, Yao Zhao, Yaofeng Sun, Yaohui Wang, Yi~Yu, Yichao Zhang, Yifan Shi, Yiliang Xiong, Ying He, Yishi Piao, Yisong Wang, Yixuan Tan, Yiyang Ma, Yiyuan Liu, Yongqiang Guo, Yuan Ou, Yuduan Wang, Yue Gong, Yuheng Zou, Yujia He, Yunfan Xiong, Yuxiang Luo, Yuxiang You, Yuxuan Liu, Yuyang Zhou, Y.~X. Zhu,
  Yanhong Xu, Yanping Huang, Yaohui Li, Yi~Zheng, Yuchen Zhu, Yunxian Ma, Ying Tang, Yukun Zha, Yuting Yan, Z.~Z. Ren, Zehui Ren, Zhangli Sha, Zhe Fu, Zhean Xu, Zhenda Xie, Zhengyan Zhang, Zhewen Hao, Zhicheng Ma, Zhigang Yan, Zhiyu Wu, Zihui Gu, Zijia Zhu, Zijun Liu, Zilin Li, Ziwei Xie, Ziyang Song, Zizheng Pan, Zhen Huang, Zhipeng Xu, Zhongyu Zhang, and Zhen Zhang.
\newblock Deepseek-r1: Incentivizing reasoning capability in llms via reinforcement learning, 2025.
\newblock \url{https://arxiv.org/abs/2501.12948}.

\bibitem[Denardo(1970)]{denardo1970linear}
Eric~V Denardo.
\newblock On linear programming in a markov decision problem.
\newblock \emph{Management Science}, 16\penalty0 (5):\penalty0 281--288, 1970.

\bibitem[Dong et~al.(2023)Dong, Xiong, Goyal, Zhang, Chow, Pan, Diao, Zhang, Shum, and Zhang]{dong2023raft}
Hanze Dong, Wei Xiong, Deepanshu Goyal, Yihan Zhang, Winnie Chow, Rui Pan, Shizhe Diao, Jipeng Zhang, Kashun Shum, and Tong Zhang.
\newblock Raft: Reward ranked finetuning for generative foundation model alignment.
\newblock \emph{arXiv preprint arXiv:2304.06767}, 2023.

\bibitem[Dubey et~al.(2024)Dubey, Jauhri, Pandey, Kadian, Al-Dahle, Letman, Mathur, Schelten, Yang, Fan, et~al.]{dubey2024llama}
Abhimanyu Dubey, Abhinav Jauhri, Abhinav Pandey, Abhishek Kadian, Ahmad Al-Dahle, Aiesha Letman, Akhil Mathur, Alan Schelten, Amy Yang, Angela Fan, et~al.
\newblock The llama 3 herd of models.
\newblock \emph{arXiv preprint arXiv:2407.21783}, 2024.

\bibitem[Flaxman et~al.(2004)Flaxman, Kalai, and McMahan]{flaxman2004online}
Abraham~D Flaxman, Adam~Tauman Kalai, and H~Brendan McMahan.
\newblock Online convex optimization in the bandit setting: gradient descent without a gradient.
\newblock \emph{arXiv preprint cs/0408007}, 2004.

\bibitem[Foerster et~al.(2018)Foerster, Farquhar, Al-Shedivat, Rockt{\"a}schel, Xing, and Whiteson]{foerster2018dice}
Jakob Foerster, Gregory Farquhar, Maruan Al-Shedivat, Tim Rockt{\"a}schel, Eric Xing, and Shimon Whiteson.
\newblock Dice: The infinitely differentiable monte carlo estimator.
\newblock In \emph{International Conference on Machine Learning}, pages 1529--1538. PMLR, 2018.

\bibitem[Garc{\i}a and Fern{\'a}ndez(2015)]{garcia2015comprehensive}
Javier Garc{\i}a and Fernando Fern{\'a}ndez.
\newblock A comprehensive survey on safe reinforcement learning.
\newblock \emph{Journal of Machine Learning Research}, 16\penalty0 (1):\penalty0 1437--1480, 2015.

\bibitem[{Gurobi Optimization, LLC}(2024)]{gurobi}
{Gurobi Optimization, LLC}.
\newblock {Gurobi Optimizer Reference Manual}, 2024.
\newblock \url{https://www.gurobi.com}.

\bibitem[Haarnoja et~al.(2017)Haarnoja, Tang, Abbeel, and Levine]{haarnoja2017reinforcement}
Tuomas Haarnoja, Haoran Tang, Pieter Abbeel, and Sergey Levine.
\newblock Reinforcement learning with deep energy-based policies.
\newblock In \emph{International conference on machine learning}, pages 1352--1361. PMLR, 2017.

\bibitem[Haarnoja et~al.(2018)Haarnoja, Zhou, Abbeel, and Levine]{haarnoja2018soft}
Tuomas Haarnoja, Aurick Zhou, Pieter Abbeel, and Sergey Levine.
\newblock Soft actor-critic: Off-policy maximum entropy deep reinforcement learning with a stochastic actor.
\newblock In \emph{International conference on machine learning}, pages 1861--1870. PMLR, 2018.

\bibitem[Hazan et~al.(2016)]{hazan2016introduction}
Elad Hazan et~al.
\newblock Introduction to online convex optimization.
\newblock \emph{Foundations and Trends{\textregistered} in Optimization}, 2\penalty0 (3-4):\penalty0 157--325, 2016.

\bibitem[Henderson et~al.(2020)Henderson, Hu, Romoff, Brunskill, Jurafsky, and Pineau]{henderson2020towards}
Peter Henderson, Jieru Hu, Joshua Romoff, Emma Brunskill, Dan Jurafsky, and Joelle Pineau.
\newblock Towards the systematic reporting of the energy and carbon footprints of machine learning.
\newblock \emph{Journal of Machine Learning Research}, 21\penalty0 (248):\penalty0 1--43, 2020.

\bibitem[Hendrycks et~al.(2021{\natexlab{a}})Hendrycks, Basart, Kadavath, Mazeika, Arora, Guo, Burns, Puranik, He, Song, et~al.]{hendrycks2021measuringcode}
Dan Hendrycks, Steven Basart, Saurav Kadavath, Mantas Mazeika, Akul Arora, Ethan Guo, Collin Burns, Samir Puranik, Horace He, Dawn Song, et~al.
\newblock Measuring coding challenge competence with apps.
\newblock \emph{arXiv preprint arXiv:2105.09938}, 2021{\natexlab{a}}.

\bibitem[Hendrycks et~al.(2021{\natexlab{b}})Hendrycks, Burns, Kadavath, Arora, Basart, Tang, Song, and Steinhardt]{hendrycks2021measuring}
Dan Hendrycks, Collin Burns, Saurav Kadavath, Akul Arora, Steven Basart, Eric Tang, Dawn Song, and Jacob Steinhardt.
\newblock Measuring mathematical problem solving with the math dataset.
\newblock \emph{arXiv preprint arXiv:2103.03874}, 2021{\natexlab{b}}.

\bibitem[Huang et~al.(2023)Huang, Chen, Mishra, Zheng, Yu, Song, and Zhou]{huang2023large}
Jie Huang, Xinyun Chen, Swaroop Mishra, Huaixiu~Steven Zheng, Adams~Wei Yu, Xinying Song, and Denny Zhou.
\newblock Large language models cannot self-correct reasoning yet.
\newblock \emph{arXiv preprint arXiv:2310.01798}, 2023.

\bibitem[Ibaraki and Katoh(1988)]{ibaraki1988resource}
Toshihide Ibaraki and Naoki Katoh.
\newblock \emph{Resource allocation problems: algorithmic approaches}.
\newblock MIT press, 1988.

\bibitem[Jacobs et~al.(1991)Jacobs, Jordan, Nowlan, and Hinton]{jacobs1991adaptive}
Robert~A Jacobs, Michael~I Jordan, Steven~J Nowlan, and Geoffrey~E Hinton.
\newblock Adaptive mixtures of local experts.
\newblock \emph{Neural computation}, 3\penalty0 (1):\penalty0 79--87, 1991.

\bibitem[Jaech et~al.(2024)Jaech, Kalai, Lerer, Richardson, El-Kishky, Low, Helyar, Madry, Beutel, Carney, et~al.]{jaech2024openai}
Aaron Jaech, Adam Kalai, Adam Lerer, Adam Richardson, Ahmed El-Kishky, Aiden Low, Alec Helyar, Aleksander Madry, Alex Beutel, Alex Carney, et~al.
\newblock Openai o1 system card.
\newblock \emph{arXiv preprint arXiv:2412.16720}, 2024.

\bibitem[Jin et~al.(2024)Jin, Yu, Shu, Zhao, Hua, Meng, Zhang, and Du]{jin2024impact}
Mingyu Jin, Qinkai Yu, Dong Shu, Haiyan Zhao, Wenyue Hua, Yanda Meng, Yongfeng Zhang, and Mengnan Du.
\newblock The impact of reasoning step length on large language models.
\newblock \emph{arXiv preprint arXiv:2401.04925}, 2024.

\bibitem[Karlin(2003)]{karlin2003mathematical}
Samuel Karlin.
\newblock \emph{Mathematical methods and theory in games, programming, and economics}, volume~2.
\newblock Courier Corporation, 2003.

\bibitem[Karwowski et~al.(2023)Karwowski, Hayman, Bai, Kiendlhofer, Griffin, and Skalse]{karwowski2023goodhart}
Jacek Karwowski, Oliver Hayman, Xingjian Bai, Klaus Kiendlhofer, Charlie Griffin, and Joar Skalse.
\newblock Goodhart's law in reinforcement learning.
\newblock \emph{arXiv preprint arXiv:2310.09144}, 2023.

\bibitem[Kumar et~al.(2024)Kumar, Zhuang, Agarwal, Su, Co-Reyes, Singh, Baumli, Iqbal, Bishop, Roelofs, et~al.]{kumar2024training}
Aviral Kumar, Vincent Zhuang, Rishabh Agarwal, Yi~Su, John~D Co-Reyes, Avi Singh, Kate Baumli, Shariq Iqbal, Colton Bishop, Rebecca Roelofs, et~al.
\newblock Training language models to self-correct via reinforcement learning.
\newblock \emph{arXiv preprint arXiv:2409.12917}, 2024.

\bibitem[Lai et~al.(2024)Lai, Tian, Chen, Yang, Peng, and Jia]{lai2024step}
Xin Lai, Zhuotao Tian, Yukang Chen, Senqiao Yang, Xiangru Peng, and Jiaya Jia.
\newblock Step-dpo: Step-wise preference optimization for long-chain reasoning of llms.
\newblock \emph{arXiv preprint arXiv:2406.18629}, 2024.

\bibitem[Lepikhin et~al.(2020)Lepikhin, Lee, Xu, Chen, Firat, Huang, Krikun, Shazeer, and Chen]{lepikhin2020gshard}
Dmitry Lepikhin, HyoukJoong Lee, Yuanzhong Xu, Dehao Chen, Orhan Firat, Yanping Huang, Maxim Krikun, Noam Shazeer, and Zhifeng Chen.
\newblock Gshard: Scaling giant models with conditional computation and automatic sharding.
\newblock \emph{arXiv preprint arXiv:2006.16668}, 2020.

\bibitem[Liang et~al.(2023)Liang, He, Jiao, Wang, Wang, Wang, Yang, Shi, and Tu]{liang2023encouraging}
Tian Liang, Zhiwei He, Wenxiang Jiao, Xing Wang, Yan Wang, Rui Wang, Yujiu Yang, Shuming Shi, and Zhaopeng Tu.
\newblock Encouraging divergent thinking in large language models through multi-agent debate.
\newblock \emph{arXiv preprint arXiv:2305.19118}, 2023.

\bibitem[Lightman et~al.(2023)Lightman, Kosaraju, Burda, Edwards, Baker, Lee, Leike, Schulman, Sutskever, and Cobbe]{lightman2023let}
Hunter Lightman, Vineet Kosaraju, Yura Burda, Harri Edwards, Bowen Baker, Teddy Lee, Jan Leike, John Schulman, Ilya Sutskever, and Karl Cobbe.
\newblock Let's verify step by step.
\newblock \emph{arXiv preprint arXiv:2305.20050}, 2023.

\bibitem[Liu et~al.(2021)Liu, Raghunathan, Liang, and Finn]{liu2021decoupling}
Evan~Z Liu, Aditi Raghunathan, Percy Liang, and Chelsea Finn.
\newblock Decoupling exploration and exploitation for meta-reinforcement learning without sacrifices.
\newblock In \emph{International conference on machine learning}, pages 6925--6935. PMLR, 2021.

\bibitem[Lorraine et~al.(2020)Lorraine, Vicol, and Duvenaud]{lorraine2020optimizing}
Jonathan Lorraine, Paul Vicol, and David Duvenaud.
\newblock Optimizing millions of hyperparameters by implicit differentiation.
\newblock In \emph{International conference on artificial intelligence and statistics}, pages 1540--1552. PMLR, 2020.

\bibitem[Madaan et~al.(2024)Madaan, Tandon, Gupta, Hallinan, Gao, Wiegreffe, Alon, Dziri, Prabhumoye, Yang, et~al.]{madaan2024self}
Aman Madaan, Niket Tandon, Prakhar Gupta, Skyler Hallinan, Luyu Gao, Sarah Wiegreffe, Uri Alon, Nouha Dziri, Shrimai Prabhumoye, Yiming Yang, et~al.
\newblock Self-refine: Iterative refinement with self-feedback.
\newblock \emph{Advances in Neural Information Processing Systems}, 36, 2024.

\bibitem[Manne(1960)]{manne1960linear}
Alan~S Manne.
\newblock Linear programming and sequential decisions.
\newblock \emph{Management Science}, 6\penalty0 (3):\penalty0 259--267, 1960.

\bibitem[Montgomery and Levine(2016)]{montgomery2016guided}
William~H Montgomery and Sergey Levine.
\newblock Guided policy search via approximate mirror descent.
\newblock \emph{Advances in Neural Information Processing Systems}, 29, 2016.

\bibitem[Nachum and Dai(2020)]{nachum2020reinforcement}
Ofir Nachum and Bo~Dai.
\newblock Reinforcement learning via fenchel-rockafellar duality.
\newblock \emph{arXiv preprint arXiv:2001.01866}, 2020.

\bibitem[Nachum et~al.(2019)Nachum, Chow, Dai, and Li]{nachum2019dualdice}
Ofir Nachum, Yinlam Chow, Bo~Dai, and Lihong Li.
\newblock Dualdice: Behavior-agnostic estimation of discounted stationary distribution corrections.
\newblock \emph{Advances in neural information processing systems}, 32, 2019.

\bibitem[Ouyang et~al.(2022)Ouyang, Wu, Jiang, Almeida, Wainwright, Mishkin, Zhang, Agarwal, Slama, Ray, et~al.]{ouyang2022training}
Long Ouyang, Jeffrey Wu, Xu~Jiang, Diogo Almeida, Carroll Wainwright, Pamela Mishkin, Chong Zhang, Sandhini Agarwal, Katarina Slama, Alex Ray, et~al.
\newblock Training language models to follow instructions with human feedback.
\newblock \emph{Advances in neural information processing systems}, 35:\penalty0 27730--27744, 2022.

\bibitem[Pan et~al.(2022)Pan, Bhatia, and Steinhardt]{pan2022effects}
Alexander Pan, Kush Bhatia, and Jacob Steinhardt.
\newblock The effects of reward misspecification: Mapping and mitigating misaligned models.
\newblock \emph{arXiv preprint arXiv:2201.03544}, 2022.

\bibitem[Pathak et~al.(2017)Pathak, Agrawal, Efros, and Darrell]{pathak2017curiosity}
Deepak Pathak, Pulkit Agrawal, Alexei~A Efros, and Trevor Darrell.
\newblock Curiosity-driven exploration by self-supervised prediction.
\newblock In \emph{International conference on machine learning}, pages 2778--2787. PMLR, 2017.

\bibitem[Peng et~al.(2019)Peng, Kumar, Zhang, and Levine]{peng2019advantage}
Xue~Bin Peng, Aviral Kumar, Grace Zhang, and Sergey Levine.
\newblock Advantage-weighted regression: Simple and scalable off-policy reinforcement learning.
\newblock \emph{arXiv preprint arXiv:1910.00177}, 2019.

\bibitem[Peters et~al.(2010)Peters, Mulling, and Altun]{peters2010relative}
Jan Peters, Katharina Mulling, and Yasemin Altun.
\newblock Relative entropy policy search.
\newblock In \emph{Proceedings of the AAAI Conference on Artificial Intelligence}, volume~24, pages 1607--1612, 2010.

\bibitem[Pham et~al.(2023)Pham, Liu, Yang, Chen, Liu, Yuan, Plummer, Wang, and Yang]{pham2023let}
Chau Pham, Boyi Liu, Yingxiang Yang, Zhengyu Chen, Tianyi Liu, Jianbo Yuan, Bryan~A Plummer, Zhaoran Wang, and Hongxia Yang.
\newblock Let models speak ciphers: Multiagent debate through embeddings.
\newblock \emph{arXiv preprint arXiv:2310.06272}, 2023.

\bibitem[Qu et~al.(2024)Qu, Zhang, Garg, and Kumar]{qu2024recursive}
Yuxiao Qu, Tianjun Zhang, Naman Garg, and Aviral Kumar.
\newblock Recursive introspection: Teaching language model agents how to self-improve.
\newblock \emph{arXiv preprint arXiv:2407.18219}, 2024.

\bibitem[Rafailov et~al.(2024)Rafailov, Sharma, Mitchell, Manning, Ermon, and Finn]{rafailov2024direct}
Rafael Rafailov, Archit Sharma, Eric Mitchell, Christopher~D Manning, Stefano Ermon, and Chelsea Finn.
\newblock Direct preference optimization: Your language model is secretly a reward model.
\newblock \emph{Advances in Neural Information Processing Systems}, 36, 2024.

\bibitem[Ray et~al.(2019)Ray, Achiam, and Amodei]{ray2019benchmarking}
Alex Ray, Joshua Achiam, and Dario Amodei.
\newblock Benchmarking safe exploration in deep reinforcement learning.
\newblock \emph{arXiv preprint arXiv:1910.01708}, 7\penalty0 (1):\penalty0 2, 2019.

\bibitem[Shazeer et~al.(2017)Shazeer, Mirhoseini, Maziarz, Davis, Le, Hinton, and Dean]{shazeer2017outrageously}
Noam Shazeer, Azalia Mirhoseini, Krzysztof Maziarz, Andy Davis, Quoc Le, Geoffrey Hinton, and Jeff Dean.
\newblock Outrageously large neural networks: The sparsely-gated mixture-of-experts layer.
\newblock \emph{arXiv preprint arXiv:1701.06538}, 2017.

\bibitem[Skalse et~al.(2022)Skalse, Howe, Krasheninnikov, and Krueger]{skalse2022defining}
Joar Skalse, Nikolaus Howe, Dmitrii Krasheninnikov, and David Krueger.
\newblock Defining and characterizing reward gaming.
\newblock \emph{Advances in Neural Information Processing Systems}, 35:\penalty0 9460--9471, 2022.

\bibitem[Slivkins et~al.(2019)]{slivkins2019introduction}
Aleksandrs Slivkins et~al.
\newblock Introduction to multi-armed bandits.
\newblock \emph{Foundations and Trends{\textregistered} in Machine Learning}, 12\penalty0 (1-2):\penalty0 1--286, 2019.

\bibitem[Snell et~al.(2022)Snell, Kostrikov, Su, Yang, and Levine]{snell2022offline}
Charlie Snell, Ilya Kostrikov, Yi~Su, Mengjiao Yang, and Sergey Levine.
\newblock Offline rl for natural language generation with implicit language q learning.
\newblock \emph{arXiv preprint arXiv:2206.11871}, 2022.

\bibitem[Snell et~al.(2024)Snell, Lee, Xu, and Kumar]{snell2024scaling}
Charlie Snell, Jaehoon Lee, Kelvin Xu, and Aviral Kumar.
\newblock Scaling llm test-time compute optimally can be more effective than scaling model parameters.
\newblock \emph{arXiv preprint arXiv:2408.03314}, 2024.

\bibitem[Sutton(2018)]{sutton2018reinforcement}
Richard~S Sutton.
\newblock Reinforcement learning: An introduction.
\newblock \emph{A Bradford Book}, 2018.

\bibitem[Touvron et~al.(2023)Touvron, Martin, Stone, Albert, Almahairi, Babaei, Bashlykov, Batra, Bhargava, Bhosale, et~al.]{touvron2023llama}
Hugo Touvron, Louis Martin, Kevin Stone, Peter Albert, Amjad Almahairi, Yasmine Babaei, Nikolay Bashlykov, Soumya Batra, Prajjwal Bhargava, Shruti Bhosale, et~al.
\newblock Llama 2: Open foundation and fine-tuned chat models.
\newblock \emph{arXiv preprint arXiv:2307.09288}, 2023.

\bibitem[Vazirani(1997)]{vazirani1997approximation}
Vijay~V Vazirani.
\newblock Approximation algorithms.
\newblock \emph{Georgia Inst. Tech}, 1997.

\bibitem[Virtanen et~al.(2020)Virtanen, Gommers, Oliphant, Haberland, Reddy, Cournapeau, Burovski, Peterson, Weckesser, Bright, {van der Walt}, Brett, Wilson, Millman, Mayorov, Nelson, Jones, Kern, Larson, Carey, Polat, Feng, Moore, {VanderPlas}, Laxalde, Perktold, Cimrman, Henriksen, Quintero, Harris, Archibald, Ribeiro, Pedregosa, {van Mulbregt}, and {SciPy 1.0 Contributors}]{2020SciPy-NMeth}
Pauli Virtanen, Ralf Gommers, Travis~E. Oliphant, Matt Haberland, Tyler Reddy, David Cournapeau, Evgeni Burovski, Pearu Peterson, Warren Weckesser, Jonathan Bright, St{\'e}fan~J. {van der Walt}, Matthew Brett, Joshua Wilson, K.~Jarrod Millman, Nikolay Mayorov, Andrew R.~J. Nelson, Eric Jones, Robert Kern, Eric Larson, C~J Carey, {\.I}lhan Polat, Yu~Feng, Eric~W. Moore, Jake {VanderPlas}, Denis Laxalde, Josef Perktold, Robert Cimrman, Ian Henriksen, E.~A. Quintero, Charles~R. Harris, Anne~M. Archibald, Ant{\^o}nio~H. Ribeiro, Fabian Pedregosa, Paul {van Mulbregt}, and {SciPy 1.0 Contributors}.
\newblock {{SciPy} 1.0: Fundamental Algorithms for Scientific Computing in Python}.
\newblock \emph{Nature Methods}, 17:\penalty0 261--272, 2020.
\newblock \doi{10.1038/s41592-019-0686-2}.

\bibitem[Wang et~al.(2024)Wang, Feng, Li, Yuan, Zhang, Pan, Wang, Hu, and Li]{wang2024make}
Xinglin Wang, Shaoxiong Feng, Yiwei Li, Peiwen Yuan, Yueqi Zhang, Boyuan Pan, Heda Wang, Yao Hu, and Kan Li.
\newblock Make every penny count: Difficulty-adaptive self-consistency for cost-efficient reasoning.
\newblock \emph{arXiv preprint arXiv:2408.13457}, 2024.

\bibitem[Wang et~al.(2022)Wang, Wei, Schuurmans, Le, Chi, Narang, Chowdhery, and Zhou]{wang2022self}
Xuezhi Wang, Jason Wei, Dale Schuurmans, Quoc Le, Ed~Chi, Sharan Narang, Aakanksha Chowdhery, and Denny Zhou.
\newblock Self-consistency improves chain of thought reasoning in language models.
\newblock \emph{arXiv preprint arXiv:2203.11171}, 2022.

\bibitem[Wei et~al.(2022)Wei, Wang, Schuurmans, Bosma, Xia, Chi, Le, Zhou, et~al.]{wei2022chain}
Jason Wei, Xuezhi Wang, Dale Schuurmans, Maarten Bosma, Fei Xia, Ed~Chi, Quoc~V Le, Denny Zhou, et~al.
\newblock Chain-of-thought prompting elicits reasoning in large language models.
\newblock \emph{Advances in neural information processing systems}, 35:\penalty0 24824--24837, 2022.

\bibitem[Wei et~al.(2024)Wei, Zhu, Zhao, Cheng, Li, L{\"u}, Cheng, Zhang, Zhang, Zeng, et~al.]{wei2024skywork}
Tianwen Wei, Bo~Zhu, Liang Zhao, Cheng Cheng, Biye Li, Weiwei L{\"u}, Peng Cheng, Jianhao Zhang, Xiaoyu Zhang, Liang Zeng, et~al.
\newblock Skywork-moe: A deep dive into training techniques for mixture-of-experts language models.
\newblock \emph{arXiv preprint arXiv:2406.06563}, 2024.

\bibitem[Xu et~al.(2024{\natexlab{a}})Xu, Li, Sun, and Qian]{xu2024adaption}
Mayi Xu, Yongqi Li, Ke~Sun, and Tieyun Qian.
\newblock Adaption-of-thought: Learning question difficulty improves large language models for reasoning.
\newblock In \emph{Proceedings of the 2024 Conference on Empirical Methods in Natural Language Processing}, pages 5468--5495, 2024{\natexlab{a}}.

\bibitem[Xu et~al.(2024{\natexlab{b}})Xu, Helenowski, Sankararaman, Jin, Peng, Han, Nie, Zhu, Zhang, Zhou, et~al.]{xu2024perfect}
Tengyu Xu, Eryk Helenowski, Karthik~Abinav Sankararaman, Di~Jin, Kaiyan Peng, Eric Han, Shaoliang Nie, Chen Zhu, Hejia Zhang, Wenxuan Zhou, et~al.
\newblock The perfect blend: Redefining rlhf with mixture of judges.
\newblock \emph{arXiv preprint arXiv:2409.20370}, 2024{\natexlab{b}}.

\bibitem[Yan et~al.(2024)Yan, Jiang, Liu, Cao, Xu, Cai, Shao, et~al.]{yan2024s}
Yuchen Yan, Jin Jiang, Yang Liu, Yixin Cao, Xin Xu, Xunliang Cai, Jian Shao, et~al.
\newblock S$^3$c-math: Spontaneous step-level self-correction makes large language models better mathematical reasoners.
\newblock \emph{arXiv preprint arXiv:2409.01524}, 2024.

\bibitem[Yang et~al.(2020)Yang, Rosca, Narasimhan, and Ramadge]{yang2020projection}
Tsung-Yen Yang, Justinian Rosca, Karthik Narasimhan, and Peter~J Ramadge.
\newblock Projection-based constrained policy optimization.
\newblock \emph{arXiv preprint arXiv:2010.03152}, 2020.

\bibitem[Yu et~al.(2023)Yu, Tao, Chen, Sun, and Yang]{yu2023beta}
Zishun Yu, Yunzhe Tao, Liyu Chen, Tao Sun, and Hongxia Yang.
\newblock $\mathcal{B}$-coder: On value-based deep reinforcement learning for program synthesis.
\newblock In \emph{The Twelfth International Conference on Learning Representations}, 2023.

\bibitem[Zelikman et~al.(2022)Zelikman, Wu, Mu, and Goodman]{zelikman2022star}
Eric Zelikman, Yuhuai Wu, Jesse Mu, and Noah Goodman.
\newblock Star: Bootstrapping reasoning with reasoning.
\newblock \emph{Advances in Neural Information Processing Systems}, 35:\penalty0 15476--15488, 2022.

\bibitem[Zemel et~al.(2013)Zemel, Wu, Swersky, Pitassi, and Dwork]{zemel2013learning}
Rich Zemel, Yu~Wu, Kevin Swersky, Toni Pitassi, and Cynthia Dwork.
\newblock Learning fair representations.
\newblock In \emph{International conference on machine learning}, pages 325--333. PMLR, 2013.

\bibitem[Zhang et~al.(2024)Zhang, Khanduri, Tsaknakis, Yao, Hong, and Liu]{zhang2024introduction}
Yihua Zhang, Prashant Khanduri, Ioannis Tsaknakis, Yuguang Yao, Mingyi Hong, and Sijia Liu.
\newblock An introduction to bilevel optimization: Foundations and applications in signal processing and machine learning.
\newblock \emph{IEEE Signal Processing Magazine}, 41\penalty0 (1):\penalty0 38--59, 2024.

\bibitem[Zhang et~al.(2020)Zhang, Vuong, and Ross]{zhang2020first}
Yiming Zhang, Quan Vuong, and Keith Ross.
\newblock First order constrained optimization in policy space.
\newblock \emph{Advances in Neural Information Processing Systems}, 33:\penalty0 15338--15349, 2020.

\bibitem[Ziebart(2010)]{ziebart2010modeling}
Brian~D Ziebart.
\newblock \emph{Modeling purposeful adaptive behavior with the principle of maximum causal entropy}.
\newblock Carnegie Mellon University, 2010.

\bibitem[Ziegler et~al.(2019)Ziegler, Stiennon, Wu, Brown, Radford, Amodei, Christiano, and Irving]{ziegler2019fine}
Daniel~M Ziegler, Nisan Stiennon, Jeffrey Wu, Tom~B Brown, Alec Radford, Dario Amodei, Paul Christiano, and Geoffrey Irving.
\newblock Fine-tuning language models from human preferences.
\newblock \emph{arXiv preprint arXiv:1909.08593}, 2019.

\bibitem[Zinkevich(2003)]{zinkevich2003online}
Martin Zinkevich.
\newblock Online convex programming and generalized infinitesimal gradient ascent.
\newblock In \emph{Proceedings of the 20th international conference on machine learning (icml-03)}, pages 928--936, 2003.

\end{thebibliography}
